\documentclass[10pt,twocolumn,letterpaper]{article}

\usepackage[pagenumbers]{cvpr} % To force page numbers, e.g. for an arXiv version

\definecolor{cvprblue}{rgb}{0.21,0.49,0.74}
\usepackage[pagebackref,breaklinks,colorlinks,allcolors=cvprblue]{hyperref}
\usepackage{amsmath}
\usepackage{bm}
\usepackage{amssymb}
\usepackage{multirow}
\usepackage{ulem} 
\usepackage{array}
\usepackage{ragged2e}
\usepackage{setspace}
\usepackage[linesnumbered,ruled,vlined]{algorithm2e}
\usepackage{colortbl}
\usepackage{tabularx}
\usepackage{algorithmic}
\usepackage{graphicx} 
\usepackage{subcaption} 
\usepackage{svg}
\usepackage{ulem} 
\usepackage{microtype}
\usepackage[accsupp]{axessibility}  % Improves PDF readability for those with disabilities.

\makeatletter
\renewcommand{\maketag@@@}[1]
{\hbox{\m@th\normalsize\normalfont#1}}%
\makeatother

\def\paperID{1926} % *** Enter the Paper ID here
\def\confName{CVPR}
\def\confYear{2025}

\title{Mind the Gap: Confidence Discrepancy Can Guide Federated Semi-Supervised Learning Across Pseudo-Mismatch}

\author{
    Yijie Liu$^{1,2}$ \quad
    Xinyi Shang$^{3}$ \quad
    Yiqun Zhang$^{4}$ \quad
    Yang Lu$^{1,2}$\thanks{Corresponding Author: Yang Lu (luyang@xmu.edu.cn)} \quad
    Chen Gong$^{5}$ \quad \\
    Jing-Hao Xue$^{3}$ \quad
    Hanzi Wang$^{1,2}$ \\
    \footnotesize{$^1$Key Laboratory of Multimedia Trusted Perception and Efficient Computing, Ministry of Education of China, Xiamen University, Xiamen, China} \\
    \footnotesize{$^2$Fujian Key Laboratory of Sensing and Computing for Smart City, School of Informatics, Xiamen University, Xiamen, China}\\
    \footnotesize{$^3$Department of Statistical Science, University College London, United Kingdom} \\ 
    \footnotesize{$^4$School of Computer Science and Technology, Guangdong University of Technology, Guangzhou, China} \\
    \footnotesize{$^5$Department of Automation, Shanghai Jiao Tong University, China} \\
    {\tt\footnotesize yijieliu@stu.xmu.edu.cn, xinyi.shang.23@ucl.ac.uk, yqzhang@gdut.edu.cn, luyang@xmu.edu.cn,}  \\
    {\tt\footnotesize chen.gong@sjtu.edu.cn, jinghao.xue@ucl.ac.uk, hanzi.wang@xmu.edu.cn}
}

\begin{document}
\maketitle

\begin{abstract}
Federated Semi-Supervised Learning (FSSL) aims to leverage unlabeled data across clients with limited labeled data to train a global model with strong generalization ability. Most FSSL methods rely on consistency regularization with pseudo-labels, converting predictions from local or global models into hard pseudo-labels as supervisory signals. However, we discover that the quality of pseudo-label is largely deteriorated by data heterogeneity, an intrinsic facet of federated learning. In this paper, we study the problem of FSSL in-depth and show that (1) heterogeneity exacerbates pseudo-label mismatches, further degrading model performance and convergence, and (2) local and global models' predictive tendencies diverge as heterogeneity increases. Motivated by these findings, we propose a simple and effective method called \textbf{S}emi-supervised \textbf{A}ggregation for \textbf{G}lobally-Enhanced \textbf{E}nsemble (SAGE), that can flexibly correct pseudo-labels based on confidence discrepancies. This strategy effectively mitigates performance degradation caused by incorrect pseudo-labels and enhances consensus between local and global models. Experimental results demonstrate that SAGE outperforms existing FSSL methods in both performance and convergence. Our code is available at \href{https://github.com/Jay-Codeman/SAGE}{https://github.com/Jay-Codeman/SAGE}.
\end{abstract}

\begin{figure}[t]
  \centering
  \begin{subfigure}{0.49\linewidth}
    \centering
    \includegraphics[width=\linewidth]{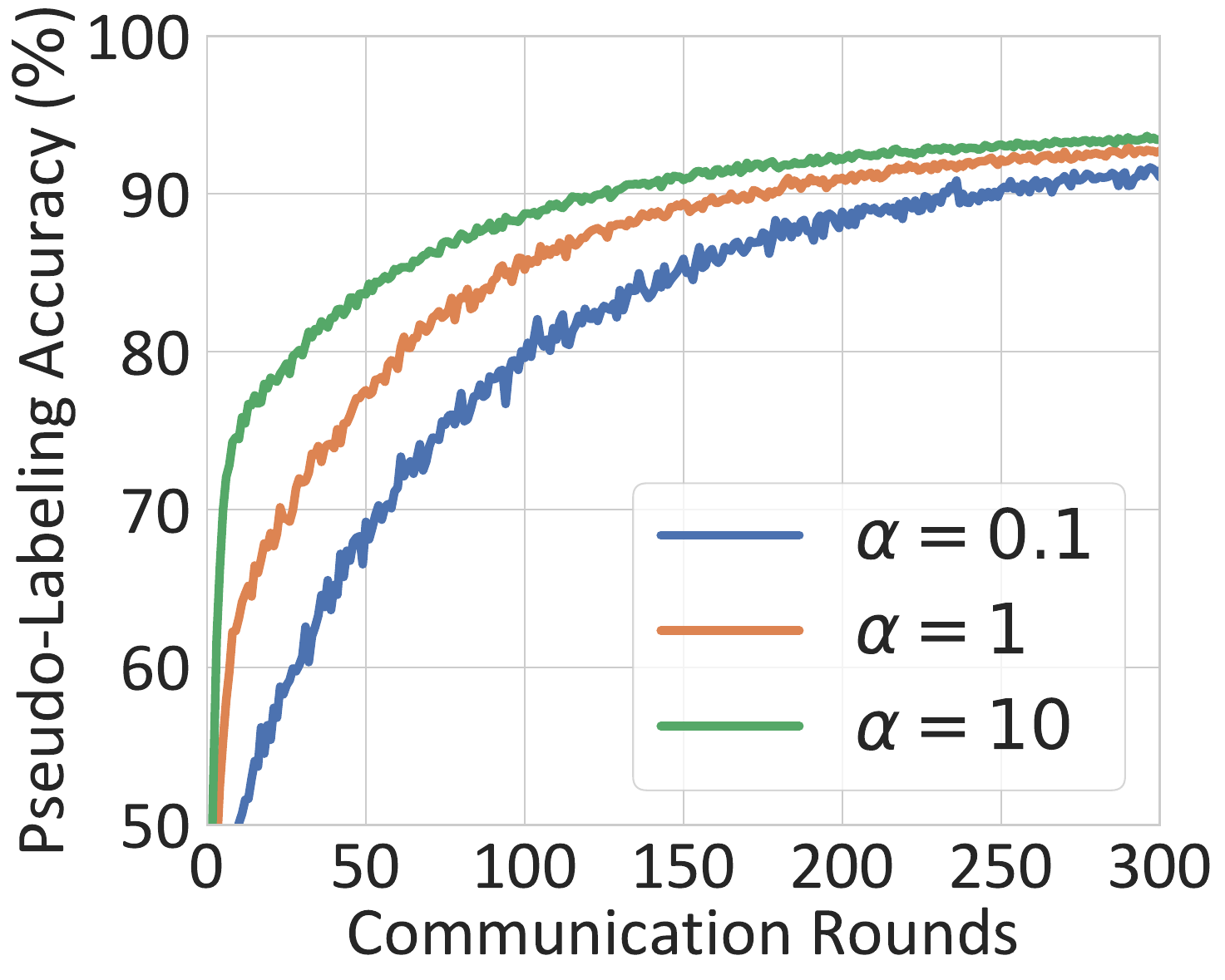}
    \caption{Pseudo-labeling accuracy.}
    \label{fig:figure1a}
  \end{subfigure}
  \hfill
  \begin{subfigure}{0.49\linewidth}
    \centering
    \includegraphics[width=\linewidth]{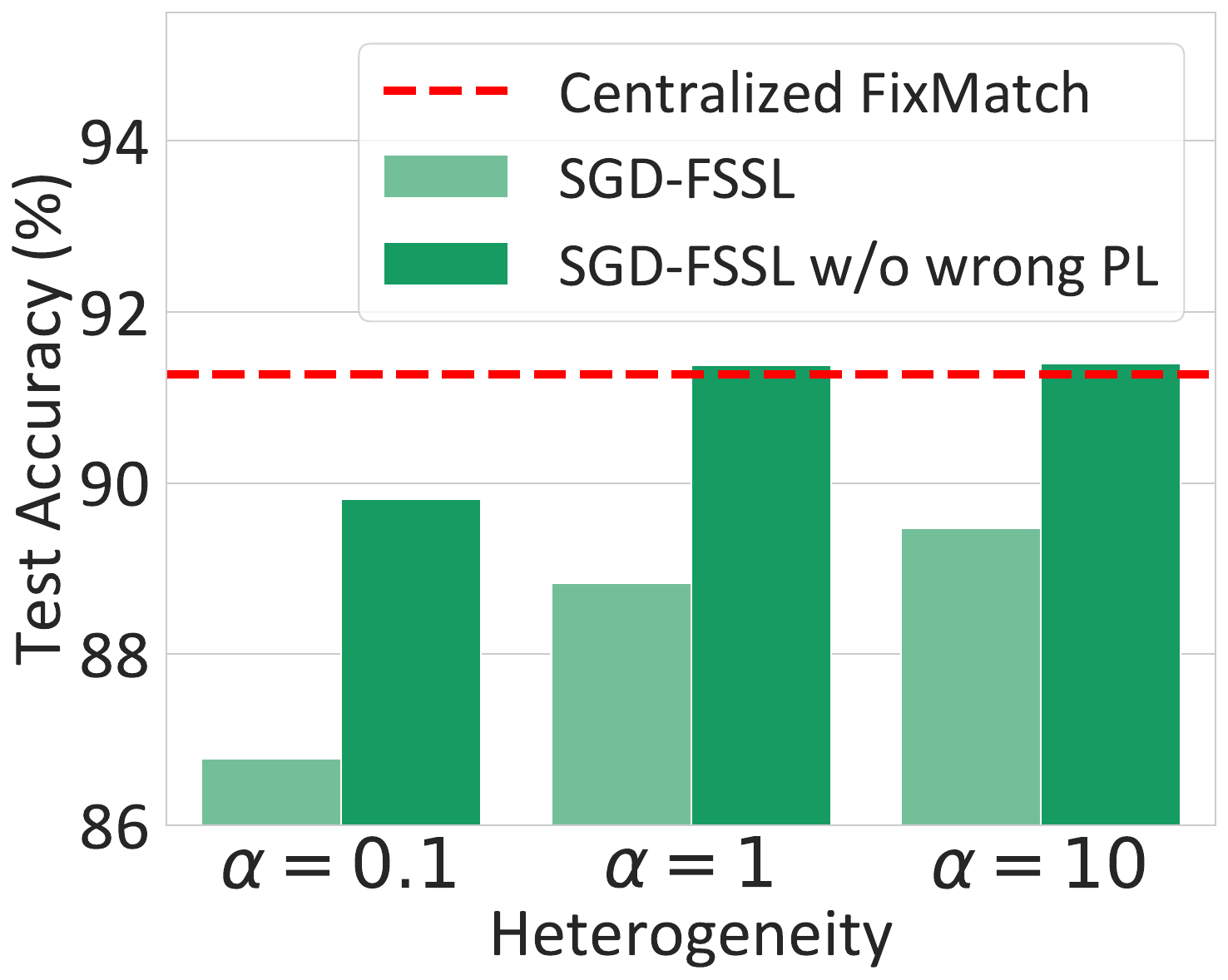}
    \caption{Test accuracy.}
    \label{fig:figure1b}
  \end{subfigure}
  \caption{\textbf{Pseudo-labeling accuracy and test accuracy under varying levels of heterogeneity }(smaller $\alpha$ indicates greater heterogeneity). In each communication round, all clients are trained using FedSGD~\cite{mcmahan2017communication} for one local epoch. From (a), we observe that as heterogeneity increases, pseudo-labeling accuracy declines. In (b), the performance gap between SGD-FSSL and Centralized FixMatch indicates the degradation caused by heterogeneity. We observe that when incorrect pseudo-labels are removed, SGD-FSSL can reach the level of centralized performance. \textbf{In short, (a) and (b)  show that data heterogeneity can negatively impact both model convergence and final test performance.}}
  \label{fig:figure1}
\end{figure}

\section{Introduction}
\label{sec:intro}

The rapid proliferation of mobile devices and the Internet of Things (IoT) has led to unprecedented growth in distributed data~\cite{lim2020federated, hoofnagle2019european}. This shift has created a pressing need for approaches that can leverage decentralized data while preserving user privacy. Federated Learning (FL) addresses this need by enabling collaborative model training directly on edge devices, sharing only model updates rather than raw data~\cite{mcmahan2017communication, konevcny2016federated}. Clients participating in FL typically possess some labeled data and conduct supervised training locally. However, when labeling costs are constrained, only a very small portion of their data may be labeled~\cite{jin2020towards}. To handle this situation, Federated Semi-Supervised Learning (FSSL) has emerged~\cite{jeong2021federated, liang2022rscfed}, allowing clients to perform Semi-Supervised Learning (SSL) on private data, leveraging a large amount of unlabeled data to improve the performance of the global model. Current research assumes data heterogeneity both within and across clients, suggesting that data distributions between clients are different (external imbalance), and within each client, labeled and unlabeled data may differ in distribution (internal imbalance)~\cite{cho2023local, bai2024combating, zhu2024estimating}. In this context, biased labels fail to generalize effectively to unseen unlabeled data.

Existing FSSL methods~\cite{diao2022semifl,li2023class, zhang2023towards, zhang2023non} typically employ consistency regularization algorithms based on pseudo-labeling, using high-confidence predictions from local or global models as pseudo-labels for unlabeled data. However, it cannot completely avoid pseudo-label mismatches even in centralized environments due to the bias of self-training~\cite{chen2022debiased}. This inspires us to explore the following questions: \textit{(1) Does heterogeneity exacerbate mismatches of hard pseudo-labels? (2) What extent do incorrect hard pseudo-labels affect FSSL model performance?}

To quantify the impact of incorrect pseudo-labels on model performance, we conduct quick experiments under varying levels of data heterogeneity. As shown in Fig.~\ref{fig:figure1}(a), with the increase of data heterogeneity (by the value of $\alpha$), the accuracy of pseudo-labels under SGD-FSSL (FedSGD+FixMatch) significantly declines with a slower convergence rate, exhibiting a clear deteriorating trend. Fortunately, as shown in Fig.~\ref{fig:figure1}(b), SGD-FSSL's accuracy improves substantially once incorrect pseudo-labels are manually removed, approaching the level of centralized FixMatch. These observations suggest that hard pseudo-labels act as aggressive supervisory signals, and their negative impact becomes especially embodied under a high level of data heterogeneity. While it is unrealistic to directly eliminate these incorrect pseudo-labels, we could consider moderately correcting them and thus mitigate their harmful effects as much as possible.

To address the problem of FSSL with the above findings, we propose a new FSSL approach called \textbf{SAGE} (\textbf{S}emi-supervised \textbf{A}ggregation for \textbf{G}lobally-enhanced \textbf{E}nsemble) to handle the scenario where the clients hold partially labeled data, apply flexible pseudo-label corrections based on the confidence perspective of the global model to mitigate the effect of erroneous hard pseudo-label signals. Firstly, we introduce a collaborative pseudo-label generation mechanism. This approach leverages the global model to guide each client, employing global distribution awareness to compensate for the scarcity of pseudo-labels in local minority classes. Secondly, we propose a dynamic, confidence-driven pseudo-label correction mechanism, inspired by an intriguing observation: as heterogeneity increases, the confidence discrepancy between local and global models gradually widens. Accordingly, we adjust the contributions of local and global hard pseudo-labels to the final pseudo-label based on their confidence discrepancies. This mechanism mitigates the impact of potentially incorrect hard pseudo-labels. Experiments show that SAGE can significantly improve the performance and convergence of the FSSL model. 

The main contributions of this paper are as follows:
\begin{itemize}
    \item This paper reveals an intriguing phenomenon: in FSSL, greater data heterogeneity results in a larger confidence discrepancy between the pseudo-labels generated by local and global models. Accordingly, we offer an explanation for the dynamic relationship between data heterogeneity and confidence discrepancies during training.
    \item We propose an FSSL method, {SAGE}, that can evaluate and flexibly correct the pseudo-labels generated by local and global models based on their confidence discrepancies under different levels of data heterogeneity, alleviating the negative impact of aggressive hard pseudo-labeling strategies.
    \item SAGE outperforms existing FSSL methods in performance and convergence across multiple datasets, demonstrating robustness under varying heterogeneous distributions. Additionally, SAGE can serve as a plugin to enhance the performance of existing FSSL methods.
\end{itemize}

\section{Related Work}
\label{sec:formatting}

\subsection{Non-IID in Federated Learning}
FL is a distributed machine learning approach enabling collaborative training across clients without sharing raw data~\cite{konevcny2016federated, mcmahan2017communication, kairouz2019advances}. Non-IID data presents a major challenge in FL, as differences in data distributions across clients significantly impact the training of federated models~\cite{zhao2018federated, li2020federated, zhu2021federated}. Numerous studies have explored the mechanisms underlying heterogeneous data’s effect and proposed solutions like classifier calibration~\cite{luo2021no, li2023no} and client selection~\cite{tang2022fedcor, chen2024heterogeneity} to alleviate performance degradation. For example, FedDECORR~\cite{shi2022towards} addresses dimensional collapse in FL due to non-IID data by regularizing local models; FedCal~\cite{peng2024fedcal} reduces global calibration error by applying client-specific calibration factors, while HiCS-FL~\cite{chen2024heterogeneity} estimates statistical heterogeneity by analyzing client updates in the output layer of the network, enabling client clustering and selection. However, these methods have yet to analyze non-IID mechanisms in federated scenarios with semi-supervised learning.

\subsection{Semi-Supervised Learning}

SSL enhances model generalization by combining limited labeled data and abundant unlabeled data, reducing dependence on labeled samples~\cite{zhou2005tri, rasmus2015semi}. Current SSL approaches fall primarily into two categories: consistency regularization and pseudo-labeling strategies. Consistency regularization~\cite{bachman2014learning, sajjadi2016regularization, xie2020unsupervised} assumes that a model should yield consistent outputs under different perturbations of the same input, using techniques like perturbation augmentation and contrastive loss to constrain the model. Pseudo-labeling strategies~\cite{lee2013pseudo, pham2021meta, hu2021simple}, meanwhile, use the model’s own predictions as labels for unlabeled samples. Recent methods like FixMatch~\cite{sohn2020fixmatch} efficiently integrate consistency regularization and pseudo-labeling through a lightweight self-training mechanism, with several studies refining this approach~\cite{zhang2021flexmatch, wang2022freematch, yang2023shrinking, lee2024cdmad}. However, pseudo-label generation in self-training methods relies heavily on prediction confidence, and in the heterogeneous setting of FL, the quality of self-generated pseudo-labels can vary greatly, making centralized SSL methods challenging to apply directly to FL scenarios.

\subsection{Federated Semi-Supervised Learning}

FSSL settings fall into three categories: (1) Label-at-Server~\cite{jeong2021federated, he2022ssfl, diao2022semifl, kim2023navigating, yang2024exploring}, where the server holds some labeled data while clients possess only unlabeled data; (2) Label-at-All-Client~\cite{jeong2021federated, zhao2022fedgan}, where each client contains a small amount of labeled data alongside a large amount of unlabeled data; and (3) Label-at-Partial-Client~\cite{liu2021federated, liang2022rscfed, li2023class, zhang2024robust, liu2024fedcd}, where only a few clients have fully labeled data, while most have only unlabeled data. Our study focuses on the Label-at-All-Client setting. Recent research~\cite{cho2023local, shang2023federated, zhang2023non, bai2024combating, zhu2024estimating} builds on FixMatch, focusing on pseudo-label selection or debiasing. However, these methods cannot avoid the impact of incorrect hard pseudo-labels in heterogeneous scenarios.

\section{Problem Formulation}
This study examines the impact of data heterogeneity on FSSL. We consider both intra-client and inter-client data heterogeneity in FSSL scenarios, with not only external imbalance across clients but also internal imbalance between labeled and unlabeled distributions within each client. Let the set of clients be $\mathcal{C} = \{C_1, C_2, \dots, C_K\}$, where each client $C_k$ trains a local model $f_{l,k}$ parameterized by $\theta_{l,k}$. During each communication round, a subset of online clients $\mathcal{C}_M \subseteq \mathcal{C}$ is randomly selected to participate in training, and the global model $f_g$ aggregates the uploaded model parameters from the clients, obtaining the global parameters $\theta_g$ as
$
\theta_g = \sum_{C_m \in \mathcal{C}_{M}} w_m \theta_{l,m},
$
where $w_m$ represents the weight for client $C_m$, determined by the proportion of its local dataset size relative to the total number of samples across all participating clients.

Each client $C_k$ maintains a private partially labeled dataset, consisting of labeled data $\mathcal{D}_k^s = \{(\mathbf{x_i}, y_i)\}_{i=1}^{N_k^s}$ and unlabeled data $\mathcal{D}_k^u = \{\mathbf{u}_i\}_{i=1}^{N_k^u}$, with $N_k^s \ll N_k^u$, both $\mathcal{D}_k^s$ and $\mathcal{D}_k^u$ demonstrate class imbalance across the label set $Y$. Specifically, there exists a significant shift between the distribution $Q_k^s(y)$ of the labeled set and the ideal uniform distribution $U=\frac{1}{|Y|}$, i.e., $\text{KL}(Q_k^s(y) \parallel U) \gg 0$. The unlabeled set $\mathcal{D}_k^u$ is assumed to follow the imbalanced distribution $Q_k^u(y)$.
For simplicity, we will omit the client index $k$ in the following sections.

\section{Proposed Method}
\subsection{Preliminary Study}
\label{section:preliminary_study}

\begin{figure}[t]
  \centering
  \begin{subfigure}{0.49\linewidth}
    \centering
    \includegraphics[width=\linewidth]{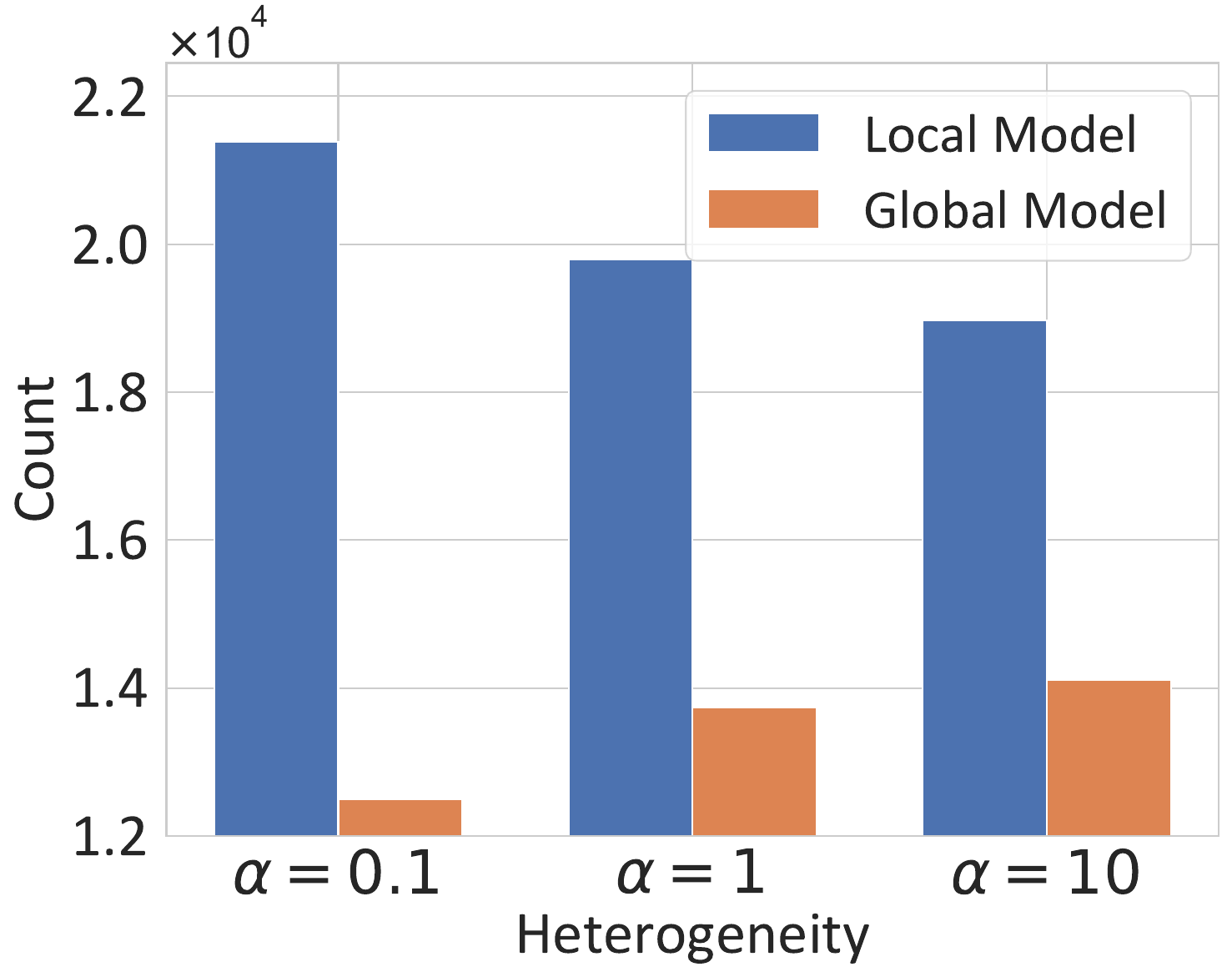}
    \caption{Pseudo-label confidence distributions of local and global models.}
    \label{fig:figure2a}
  \end{subfigure}
  \hfill
  \begin{subfigure}{0.49\linewidth}
    \centering
    \includegraphics[width=\linewidth]{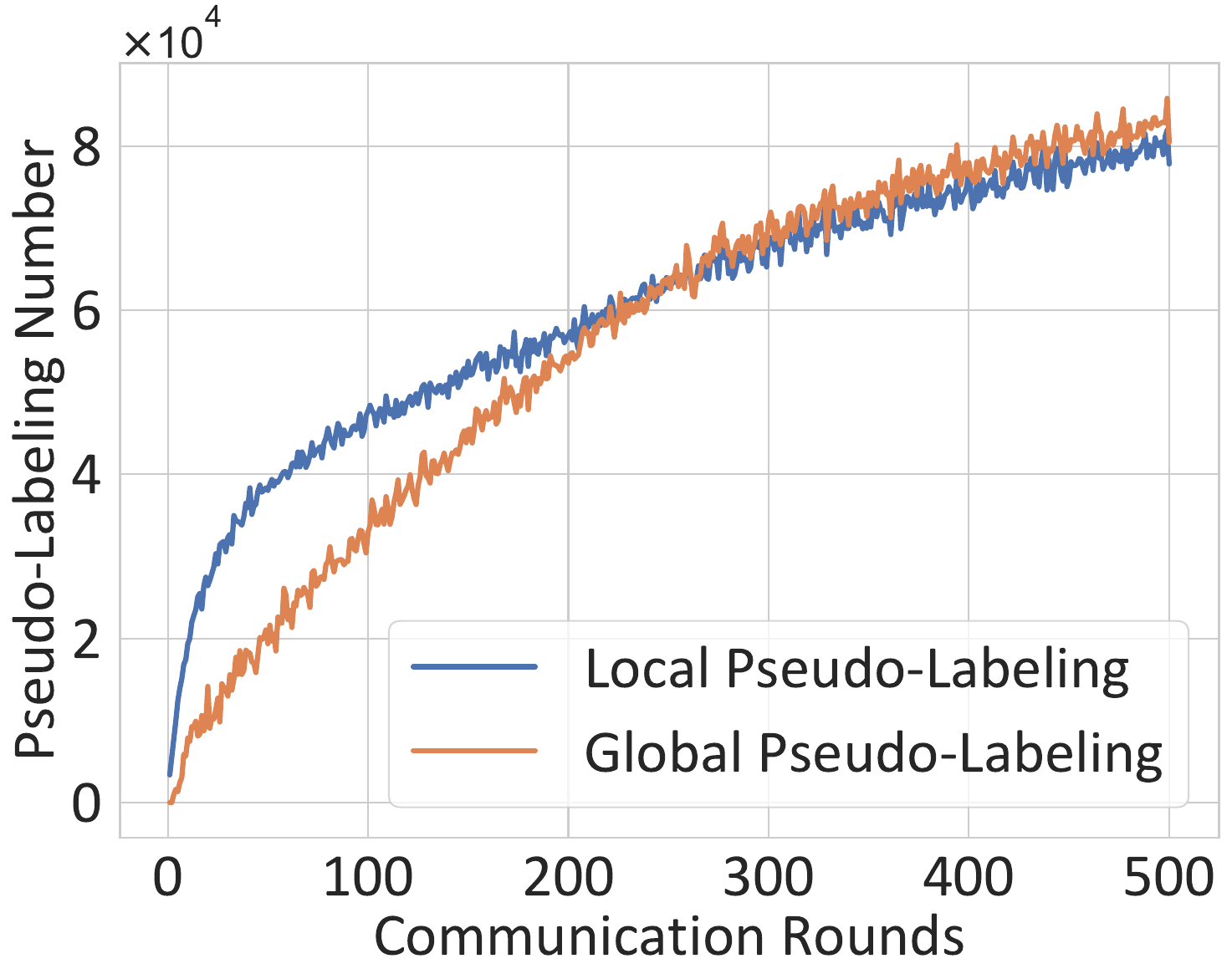}
    \caption{The number of pseudo-labels under $\alpha=0.1$.}
    \label{fig:figure2b}
  \end{subfigure}

  % \caption{The difference of the pseudo-labeling ability between local and global models on CIFAR-100. (a) shows the distributions of pseudo-labels with confidence greater than 0.99. As heterogeneity increases (with smaller $\alpha$), the local model tends to make more confident predictions, while the global model tends to make less confident predictions. The trend is also reflected in the number of pseudo-labels in (b), where the confident predictions of local model makes a higher utilization rate of unlabeled data in the early stages of training.}

    \caption{Differences of the pseudo-labeling ability between local and global models on CIFAR-100. (a) shows the distributions of pseudo-labels with confidence greater than 0.99. As heterogeneity increases (with smaller $\alpha$), the local and global models exhibit opposite trends. The difference is also reflected in the number of pseudo-labels in (b).}
  \label{fig:preliminary_study}
\end{figure}

Our goal is to moderately correct potentially incorrect pseudo-labels to mitigate their impact, with the local and global models providing two distinct perspectives on pseudo-labels. To this end, we conduct exploratory experiments to investigate the pseudo-labeling differences between local and global models as data heterogeneity increases. We analyze the confidence distribution of pseudo-labels from both models and track the number of pseudo-labels assigned throughout training. As shown in Fig.~\ref{fig:preliminary_study}(a), the confidence of the local model’s pseudo-labels shifts more toward the high-confidence region, while the global model exhibits the opposite trend. Additionally, as shown in  Fig.~\ref{fig:preliminary_study}(b), the local model assigns a higher number of pseudo-labels than the global model in the early stages of training. More exploratory experimental results are shown in Appendix~\ref{sec:additional_exploratory_experiments}. Based on these experiments, we summarize this phenomenon through two key observations:

\textbf{Observation 1.} \hypertarget{obs1}{\textit{As heterogeneity intensifies, the pseudo-label predictions of the local model grow more confident, while those of the global model become more conservative.} }

\textbf{Observation 2.} \hypertarget{obs2}{\textit{The local model exhibits a higher utilization rate of unlabeled data in the early training stages compared to the global model.}}

We further analyze the rationale behind these observations to explain why increasing heterogeneity leads to differing predictive tendencies in pseudo-labels between local and global models. The detailed derivation towards the above analysis is provided in Appendix~\ref{sec:analysis} and \ref{sec:experimental_support}. The analysis is as follows:

\textbf{Remark 1.} \hypertarget{remark1}{\textit{The entropy of the local model’s predictive distribution, $H(p(y | \mathbf{x}, \mathcal{D}^u))$, is influenced by the entropy of the prior distribution $H(p(y | \mathcal{D}^u))$ and is related to the entropy of the local data distribution $H(Q^u(y))$.} }

\begin{figure*}[t]
    \centering
    \includegraphics[width=\textwidth]{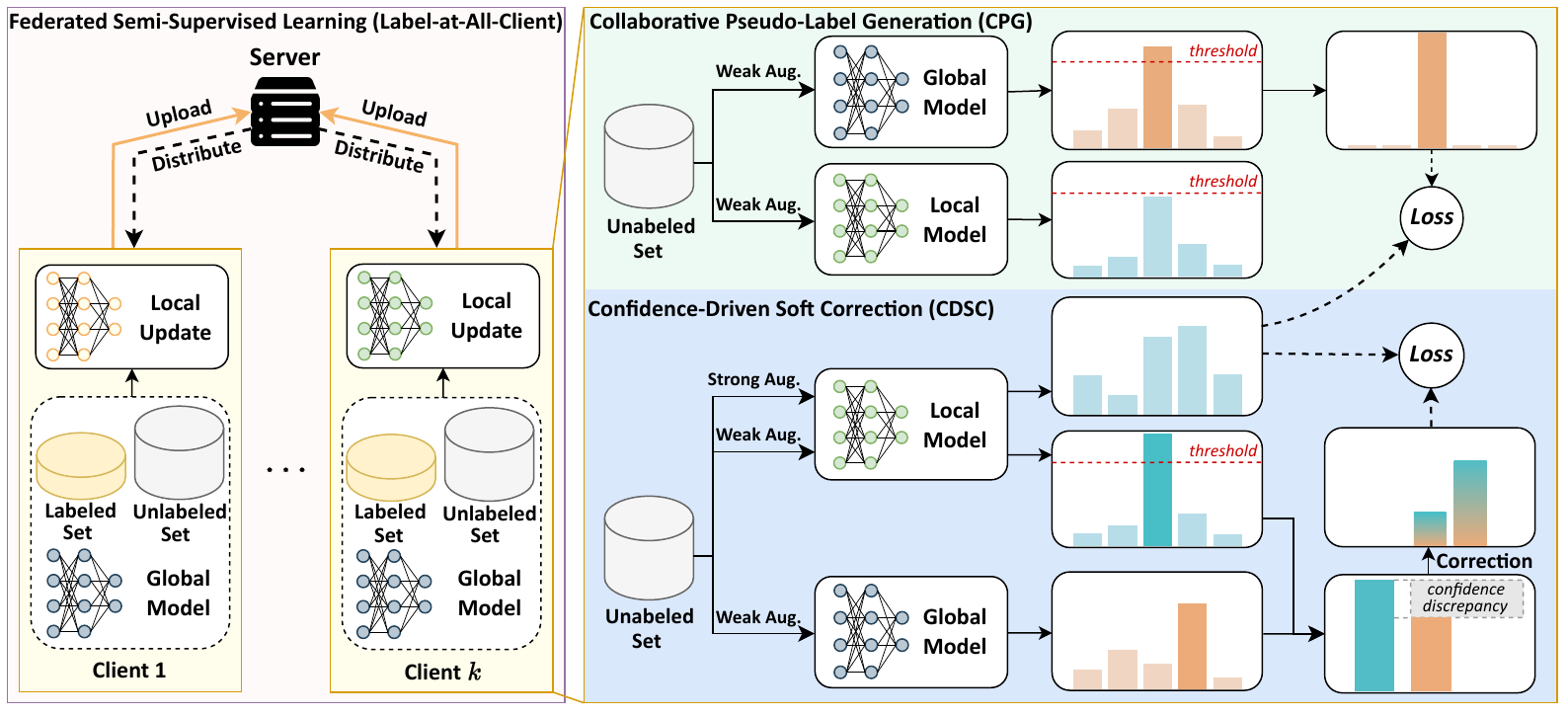}
    \caption{\textbf{Framework of the proposed SAGE.} This framework demonstrates the pseudo-labeling strategy of SAGE in the label-at-all-client scenario. The global model’s pseudo-labels provide supplementary information when the local model lacks confidence and are dynamically adjusted based on confidence discrepancies between the local and global models.}

    \label{fig:figure3}
\end{figure*}

\textbf{Remark 2.} \hypertarget{remark2}{\textit{The global model's high-confidence predictions increasingly focus on classes with higher consistency across clients, demonstrating more conservative behavior.}}

They suggest that the local model tends to overfit when faced with Non-IID data, relying excessively on its imbalanced distribution and being \textit{overly confident} in its predictions, while the global model exhibits a \textit{lack of confidence} as it attempts to create a model that can adapt to the data distribution of all clients.

Based on the above analysis, the pseudo-labeling strategies of the local and global models exhibit substantial discrepancies, offering an opportunity to mitigate the impact of potentially incorrect pseudo-labels by leveraging these discrepancies. To address this, we propose \underline{C}ollaborative \underline{P}seudo-Label \underline{G}eneration (CPG) and \underline{C}onfidence-\underline{D}riven \underline{S}oft \underline{C}orrection (CDSC), improving unsupervised data utilization while ensuring pseudo-label quality and using flexible pseudo-labels to avoid the radical impacts of hard pseudo-labels. The framework of SAGE approach is shown in Fig.~\ref{fig:figure3}.

\subsection{Collaborative Pseudo-Label Generation}

As discussed above, the pseudo-labeling abilities of the local model $f_l$ and the global model $f_g$ have their respective strengths and weaknesses: $f_l$ is trained on local data, generating a large number of pseudo-labels with high utilization of unsupervised data, but the accuracy of these pseudo-labels cannot be guaranteed. On the other hand,  $f_g$ generates fewer pseudo-labels but has a better understanding of the overall data distribution, compensating for the shortcomings of the local model. It can offer robust pseudo-label support to the local model for minority classes, thereby mitigating training errors resulting from the exclusive reliance on local pseudo-labeling strategies. Therefore, we anticipate that integrating the strengths of both models will reduce training errors caused by reliance solely on local pseudo-labels, thereby enhancing the overall pseudo-labeling accuracy.

Therefore, we propose Collaborative Pseudo-Label Generation (CPG) to ensure pseudo-labeling accuracy while enhancing the utilization of unlabeled data. For each unsupervised sample $\mathbf{u} \in \mathcal{D}^u$, we compute the weakly augmented prediction outputs of the local model and the global model, denoted as $p_l(\mathbf{u}) = f_l(\alpha(\mathbf{u}))$ and $p_g(\mathbf{u}) = f_g(\alpha(\mathbf{u}))$, where $\alpha(\mathbf{u})$ represents the weak augmentation (e.g., using only flip-and-shift data augmentation) applied to the unsupervised sample $\mathbf{u}$. We will omit $\mathbf{u}$ in the following text to avoid redundancy. We initially assign pseudo-labels based on the predictions of $f_l$ and $f_g$:
\begin{equation}
\hat{y} = 
\begin{cases}
    \arg \max(p_l) & \text{if } \max(p_l) > \tau, \\
    \arg \max(p_g) & \text{else if } \max(p_g) > \tau, \\
    \text{N/A} & \text{otherwise,}
\label{eq:collaborative_pseudo_generation}
\end{cases}
\end{equation}
where $\tau$ is the confidence threshold.
This strategy, derived from Observation~\hyperlink{obs2}{2}, prioritizes obtaining pseudo-labels from the local model and supplements them with predictions from the global model when local confidence is insufficient. This approach ensures pseudo-label quality while further enhancing the utilization of unlabeled data. Building on this, we will further correct pseudo-labels.

\subsection{Confidence-Driven Soft Correction}
CPG enables local models to maintain a high utilization rate of unlabeled data while compensating for the scarcity of pseudo-labels in local minority classes. Building on this, we further aim to utilize the conservative predictions of the global model to mitigate the impact of incorrect local pseudo-labels. From Observation~\hyperlink{obs1}{1}, we can infer that: 
\textbf{\textit{as heterogeneity intensifies, the confidence discrepancy between the local and global models widens.}} 
This insight suggests that the confidence discrepancy between the local and global model can serve as a measure of local imbalance in the predicted class. Specifically, a larger confidence difference between $f_g$ and $f_l$ indicates a greater discrepancy between the local and global distributions for the locally predicted pseudo-label class of that sample. In such cases, we assign greater weight to the global model to ensure pseudo-label robustness. Conversely, when the confidence discrepancy between $f_g$ and $f_l$ is small, it suggests that the local pseudo-label predictions are reliable. In this scenario, the local model is able to capture the characteristics of the local distribution during the current training iteration. Below, we provide a detailed explanation of the confidence-driven soft correction mechanism.

First, we calculate the confidence difference $\Delta C$ between $f_l$ and $f_g$ to characterize the discrepancy between the models:
\begin{equation}
\Delta C = |\max(p_l) - \max(p_g)|.
\end{equation}
Then, based on $\Delta C$, we dynamically adjust the contribution of each model to the pseudo-labels. We define a dynamic correction coefficient $\lambda(\cdot)$ to regulate the contribution of the local and global model pseudo-labels. As $\Delta C$ increases, we should decrease the influence of local pseudo-labels and rely more on the conservative predictions of the global model. Therefore, the correction coefficient takes the form of an exponential decay:
\begin{equation}
\textstyle
\lambda = \exp(-\kappa \cdot \Delta C),
\label{eq:Dynamic Correction Coefficient}
\end{equation}
where $\kappa$ is a hyperparameter that controls the sensitivity of the correction coefficient.

Next, based on $\lambda$, we perform linear interpolation between the predictions of $f_l$ and $f_g$. We first convert the local pseudo-label and the global predicted class into one-hot form:
\begin{align}
\delta_l &= \text{one-hot}(\arg \max(p_l)), \\
\delta_g &= \text{one-hot}(\arg \max(p_g)).
\end{align}
Then the corrected local pseudo-label is obtained through linear interpolation of them:
\begin{equation}
\textstyle
\tilde{y} = \lambda \cdot \delta_l + (1 - \lambda) \cdot \delta_g.
\label{eq:cdsc_pseudo_label}
\end{equation}
Based on this linear correction, when the confidence predictions of $f_l$ and $f_g$ are more consistent, we rely more on $f_l$'s prediction; when there is a larger discrepancy, we rely more on $f_g$'s prediction. The final pseudo-label $\hat{y}$ can be expressed as:

\begin{equation}
\hat{y} = 
\begin{cases}
   \tilde{y} & \text{if } \max(p_l) > \tau, \\
    \arg \max(p_g) & \text{else if } \max(p_g) > \tau, \\
    \text{N/A} & \text{otherwise.}
\end{cases}
\label{pseudo_correction}
\end{equation}
Through dynamic and flexible correction, CDSC mitigates the radical impact of hard pseudo-labels.

\subsection{Loss Functions}

\renewcommand{\arraystretch}{1.13}
\begin{table*}[t]
  \centering
    \caption{Experimental results on CIFAR-10, CIFAR-100, SVHN and CINIC-10 under 10\% label. Bold text indicates the best result, while underlined text indicates the second-best result. The last row presents the improvement of SAGE over existing methods.}
  \resizebox{1\textwidth}{!}{
\begin{tabular}{rlcccccccccccc}
\hline\hline
\multicolumn{2}{l}{} &
  \multicolumn{3}{c}{CIFAR-10} &
  \multicolumn{3}{c}{CIFAR-100} &
  \multicolumn{3}{c}{SVHN} &
  \multicolumn{3}{c}{CINIC-10} \\
\multicolumn{2}{c}{\multirow{-2}{*}{\textbf{Methods}}} &
  $\alpha=0.1$ &
  $\alpha=0.5$ &
  $\alpha=1$ &
  $\alpha=0.1$ &
  $\alpha=0.5$ &
  $\alpha=1$ &
  $\alpha=0.1$ &
  $\alpha=0.5$ &
  $\alpha=1$ &
  $\alpha=0.1$ &
  $\alpha=0.5$ &
  $\alpha=1$ \\ \hline\hline
\multicolumn{2}{r}{\textbf{SL methods}} &
  \multicolumn{1}{l}{} &
  \multicolumn{1}{l}{} &
  \multicolumn{1}{l}{} &
  \multicolumn{1}{l}{} &
  \multicolumn{1}{l}{} &
  \multicolumn{1}{l}{} &
  \multicolumn{1}{l}{} &
  \multicolumn{1}{l}{} &
  \multicolumn{1}{l}{} &
  \multicolumn{1}{l}{} &
  \multicolumn{1}{l}{} &
  \multicolumn{1}{l}{} \\ \hline\hline
\multicolumn{2}{r}{FedAvg} &
  69.60 &
  68.88 &
  69.39 &
  34.08 &
  33.21 &
  35.31 &
  82.40 &
  83.40 &
  78.60 &
  57.17 &
  60.09 &
  61.54 \\
\multicolumn{2}{r}{FedProx} &
  68.58 &
  69.53 &
  68.00 &
  34.20 &
  34.07 &
  34.88 &
  81.67 &
  83.77 &
  83.77 &
  58.05 &
  60.71 &
  62.82 \\
\multicolumn{2}{r}{FedAvg-SL} &
  90.46 &
  91.24 &
  91.32 &
  67.98 &
  68.83 &
  69.10 &
  94.11 &
  94.41 &
  94.40 &
  77.82 &
  80.42 &
  81.29 \\ \hline\hline
\multicolumn{2}{r}{\textbf{SSL methods}} &
  \multicolumn{1}{l}{} &
  \multicolumn{1}{l}{} &
  \multicolumn{1}{l}{} &
  \multicolumn{1}{l}{} &
  \multicolumn{1}{l}{} &
  \multicolumn{1}{l}{} &
  \multicolumn{1}{l}{} &
  \multicolumn{1}{l}{} &
  \multicolumn{1}{l}{} &
  \multicolumn{1}{l}{} &
  \multicolumn{1}{l}{} &
  \multicolumn{1}{l}{} \\ \hline\hline
\multicolumn{2}{r}{FixMatch-LPL} &
  82.98 &
  84.36 &
  84.69 &
  49.32 &
  49.67 &
  49.55 &
  89.68 &
  91.33 &
  91.91 &
  68.02 &
  70.67 &
  72.69 \\
\multicolumn{2}{r}{FixMatch-GPL} &
  84.56 &
  {\uline{86.05}} &
  86.66 &
  48.96 &
  51.80 &
  52.19 &
  90.50 &
  91.94 &
  92.31 &
  {\uline{71.67}} &
  73.26 &
  74.80 \\
\multicolumn{2}{r}{FedProx+FixMatch} &
  {\uline{84.60}} &
  85.49 &
  86.95 &
  48.42 &
  48.51 &
  49.33 &
  90.46 &
  91.36 &
  91.25 &
  68.62 &
  70.67 &
  72.69 \\
\multicolumn{2}{r}{FedAvg+FlexMatch} &
  84.21 &
  86.00 &
  86.57 &
  49.91 &
  51.39 &
  51.79 &
  52.58 &
  55.59 &
  60.50 &
  69.20 &
  71.87 &
  73.42 \\ \hline\hline
\multicolumn{2}{r}{\textbf{FSSL methods}} &
  \multicolumn{1}{l}{} &
  \multicolumn{1}{l}{} &
  \multicolumn{1}{l}{} &
  \multicolumn{1}{l}{} &
  \multicolumn{1}{l}{} &
  \multicolumn{1}{l}{} &
  \multicolumn{1}{l}{} &
  \multicolumn{1}{l}{} &
  \multicolumn{1}{l}{} &
  \multicolumn{1}{l}{} &
  \multicolumn{1}{l}{} &
  \multicolumn{1}{l}{} \\ \hline\hline
\multicolumn{2}{r}{FedMatch~\cite{jeong2021federated}} &
  75.35 &
  77.86 &
  78.00 &
  32.23 &
  31.49 &
  35.75 &
  88.63 &
  89.20 &
  89.23 &
  51.94 &
  56.27 &
  70.22 \\
\multicolumn{2}{r}{FedLabel~\cite{cho2023local}} &
  62.85 &
  79.46 &
  79.17 &
  {\uline{50.88}} &
  {\uline{52.21}} &
  {\uline{52.38}} &
  89.31 &
  91.51 &
  91.16 &
  67.64 &
  70.56 &
  72.80 \\
\multicolumn{2}{r}{FedLoke~\cite{zhang2023non}} &
  83.32 &
  82.22 &
  81.87 &
  39.29 &
  40.46 &
  39.96 &
  89.94 &
  90.00 &
  89.45 &
  59.03 &
  61.60 &
  63.21 \\
\multicolumn{2}{r}{FedDure~\cite{bai2024combating}} &
  84.60 &
  85.88 &
  87.34 &
  48.27 &
  51.09 &
  50.79 &
  {\uline{92.87}} &
  {\uline{93.49}} &
  {\uline{94.19}} &
  70.86 &
  {\uline{73.37}} &
  {\uline{74.89}} \\
\multicolumn{2}{r}{FedDB~\cite{zhu2024estimating}} &
  83.99 &
  85.28 &
  {\uline{87.49}} &
  48.43 &
  50.11 &
  51.55 &
  92.56 &
  93.00 &
  93.14 &
  69.44 &
  72.60 &
  73.61 \\ 
\multicolumn{2}{l}{} &
  \cellcolor[HTML]{EFEFEF}\textbf{87.05} &
  \cellcolor[HTML]{EFEFEF}\textbf{88.05} &
  \cellcolor[HTML]{EFEFEF}\textbf{89.08} &
  \cellcolor[HTML]{EFEFEF}\textbf{54.18} &
  \cellcolor[HTML]{EFEFEF}\textbf{55.82} &
  \cellcolor[HTML]{EFEFEF}\textbf{56.06} &
  \cellcolor[HTML]{EFEFEF}\textbf{93.85} &
  \cellcolor[HTML]{EFEFEF}\textbf{94.27} &
  \cellcolor[HTML]{EFEFEF}\textbf{94.65} &
  \cellcolor[HTML]{EFEFEF}\textbf{74.59} &
  \cellcolor[HTML]{EFEFEF}\textbf{75.74} &
  \cellcolor[HTML]{EFEFEF}\textbf{76.68} \\
\multicolumn{2}{r}{\multirow{-2}{*}{SAGE (ours)}} &
  {\color[HTML]{009901} ↑ 2.45} &
  {\color[HTML]{009901} ↑ 2.00} &
  {\color[HTML]{009901} ↑ 1.59} &
  {\color[HTML]{009901} ↑ 3.3} &
  {\color[HTML]{009901} ↑ 3.61} &
  {\color[HTML]{009901} ↑ 3.68} &
  {\color[HTML]{009901} ↑ 0.98} &
  {\color[HTML]{009901} ↑ 0.78} &
  {\color[HTML]{009901} ↑ 0.46} &
  {\color[HTML]{009901} ↑ 2.92} &
  {\color[HTML]{009901} ↑ 2.37} &
  {\color[HTML]{009901} ↑ 1.79} \\ \hline\hline
\end{tabular}
    }

  \label{table:experimental_10per}
\end{table*}

For a batch of unlabeled samples $B_u$, we use KL divergence to compute the unsupervised loss between the corrected soft pseudo-label and the local model’s strongly augmented prediction for the sample $\mathbf{u}$, denoted as $p_{l}(\mathcal{A}(\mathbf{u}))$:
\begin{equation}
L_u = \frac{1}{|B_u|} \sum_{\mathbf{u} \in B_u} \text{KL} \left( p_{l}(\mathcal{A}(\mathbf{u})) \, \Big\| \, \hat{y} \right),
\end{equation}
where $\mathcal{A}(\mathbf{u})$ refers to RandAugment with random magnitude~\cite{cubuk2020randaugment}.
% \textcolor{red}{(JHX: please add the definition of $\mathcal{A}(\mathbf{u})$.)}
For a batch of labeled samples $B_s$, we calculate the cross entropy between the local model’s predictions and the ground-truth labels: $L_s = \frac{1}{|B_s|}\sum_{\mathbf{x} \in B_s}  \mathcal{L}_{CE}(p_l(y|\mathbf{x}, \mathbf{y}))$, where $\mathcal{L}_{CE}$ is the cross-entropy loss. The final loss is a combination of supervised and unsupervised loss:
\begin{equation}
\mathcal{L} = L_s + \mu_u \cdot L_u.
\end{equation}
We follow the setup in FixMatch~\cite{sohn2020fixmatch} where $L_s$ and $L_u$ have the same weight, i.e., $\mu_u = 1$.

The process of SAGE is presented in Algorithm~\ref{algorithm:sage} in Appendix~\ref{sec:workflow}. Using the CPG and CDSC components, SAGE leverages the high utilization of the local model and the balanced distribution of the global model, enabling a ``safer" utilization of unlabeled data. This approach mitigates the harmful effects of erroneous hard pseudo-labels and enhances the consensus between local and global models.

\section{Experiments}

\subsection{Experimental Setup}
    \paragraph{Datasets.}We evaluated the SAGE method on the CIFAR-10, CIFAR-100, SVHN, and CINIC-10 datasets~\cite{krizhevsky2009learning, netzer2011reading, darlow2018cinic}. For each dataset, we divided the labeled and unlabeled datasets per class with label proportions of 10\% and 20\%. We focus on evaluating the performance of methods under more challenging conditions of heterogeneous data. In line with previous work in the FSSL field~\cite{cho2023local, bai2024combating, zhu2024estimating}, we simulated both inter-client and intra-client imbalances by sampling labeled and unlabeled data from a Dirichlet distribution $\text{Dir}(\alpha)$ and allocating them equally to each client. We simulated three levels of heterogeneity: $\alpha \in \{0.1, 0.5, 1\}$, A smaller $\alpha$ value indicates higher data heterogeneity. The specific data distribution is shown in the visualization of Fig.~\ref{fig:figure9} in Appendix~\ref{sec:class_distribution_mismatch}. For all methods, we follow the FixMatch setup and add labeled samples without labels into the unlabeled dataset to enhance sample diversity in the unsupervised dataset. We compared the following methods in our experiments:
\begin{itemize}
    \item \textbf{FL methods (FedAvg~\cite{mcmahan2017communication}, FedProx~\cite{li2020federated_fedprox}, FedAvg-SL):} For FedAvg and FedProx, models are trained via supervised federated learning using only the labeled dataset. FedAvg-SL denotes the standard federated training of FedAvg on the fully labeled dataset, indicating the ideal upper bound.

    \item \textbf{Vanilla combinations:} These methods simply combine SSL methods with FL methods. Notably, for FedAvg+FixMatch, we further subdivided it into “local model pseudo-labeling” and “global model pseudo-labeling” to illustrate differences in pseudo-labeling capabilities between local and global models, abbreviated as \textbf{FixMatch-LPL} and \textbf{FixMatch-GPL}.

    \item \textbf{FSSL methods:} SAGE is compared with state-of-the-art FSSL methods, including FedMatch~\cite{jeong2021federated}, FedLoke~\cite{zhang2023non}, FedLabel~\cite{cho2023local}, FedDure~\cite{bai2024combating}, and FedDB~\cite{zhu2024estimating}. All of them follow the Label-at-All-Client scenario.
\end{itemize}

\paragraph{Implementation Details.} We assume a total of $|\mathcal{C}|=20$ clients participating in FL, with $|\mathcal{C}_M|=8$ clients randomly selected each round for global training. ResNet-8 serves as the backbone network locally, with the number of local epochs set to $E=5$ and the local learning rate set to $\gamma=0.1$. Except for FlexMatch, the pseudo-label confidence threshold for all other methods is set to $\tau=0.95$. Unless otherwise specified, SAGE follows the FixMatch setup in this section. All experiments are conducted three times, with standard deviations shown as error bars in the figures.

\begin{figure*}[t]
    % Figure 4
    \begin{minipage}[t]{0.32\textwidth} % 
        \centering
        \includegraphics[width=\textwidth]{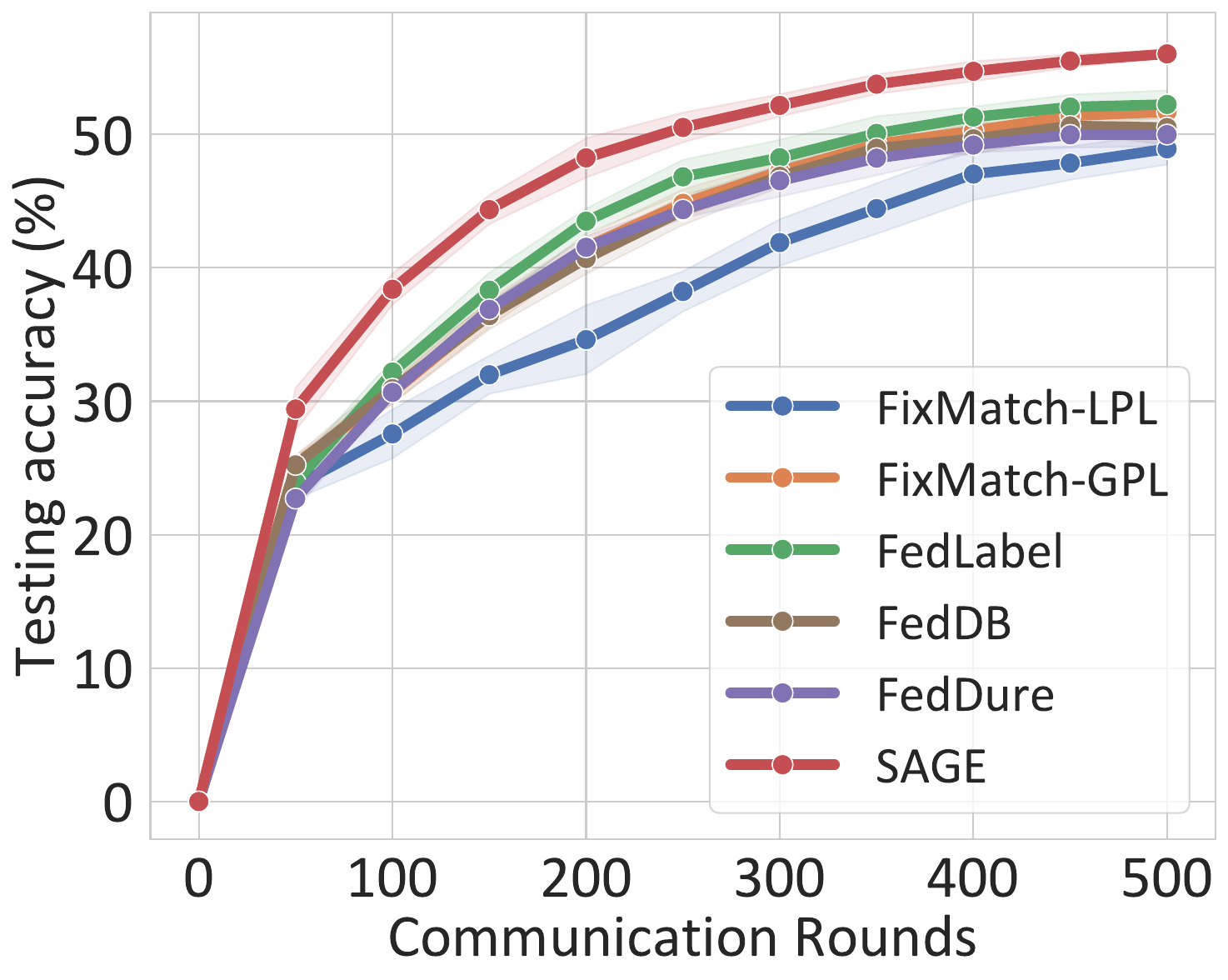}
        \caption{Convergence curves of SAGE and other baselines on CIFAR-100 with $\alpha=1$.} 
        \label{fig:figure4}
    \end{minipage}
    \hfill
    % Figure 5
    \begin{minipage}[t]{0.32\textwidth}
        \centering
        \includegraphics[width=\textwidth]{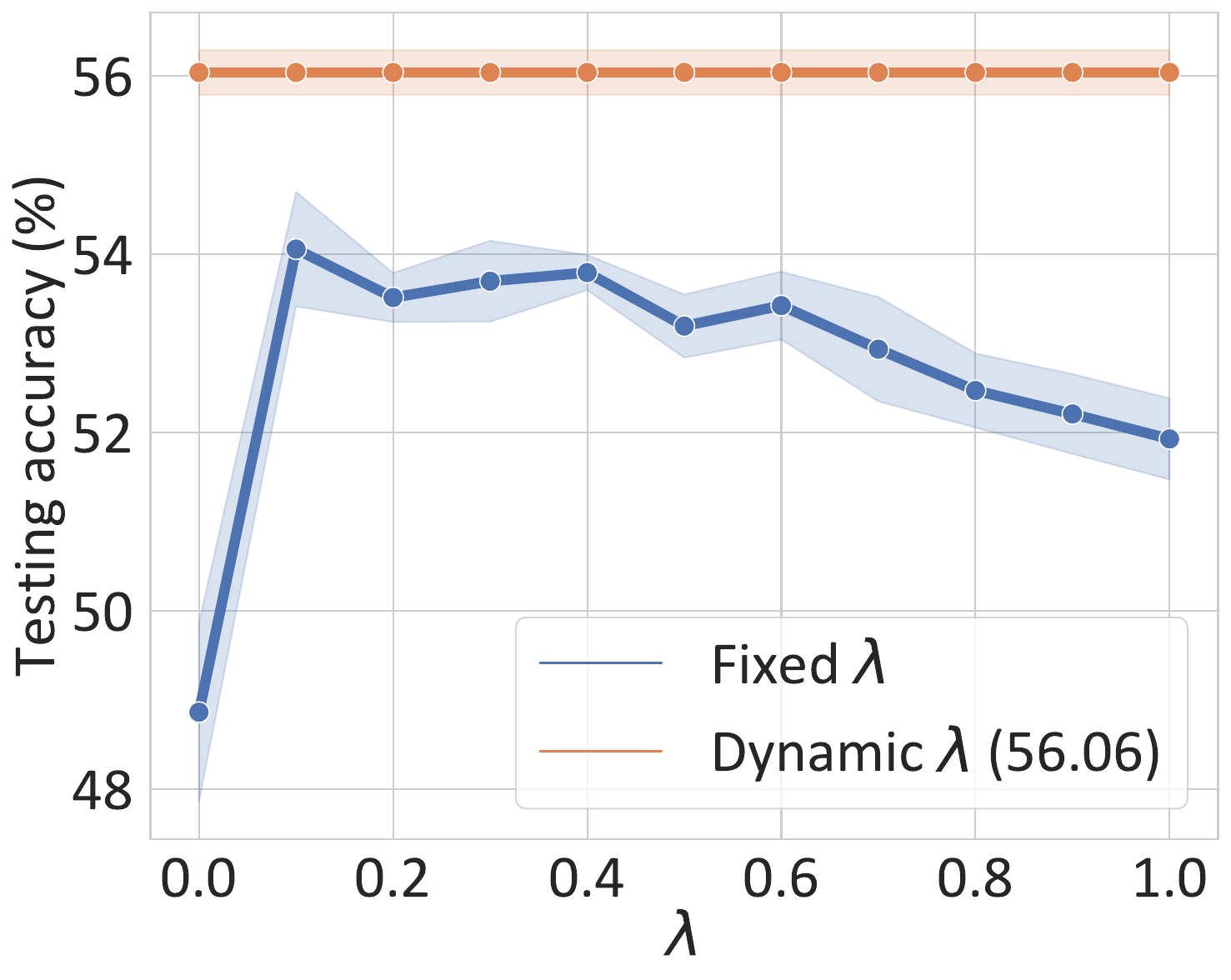}
        \caption{Ablation on Dynamic Correction Coefficient $\lambda$.} 
        \label{fig:figure5}
    \end{minipage}
    \hfill
    % Figure 7
    \begin{minipage}[t]{0.32\textwidth}
        \centering
        \includegraphics[width=\textwidth]{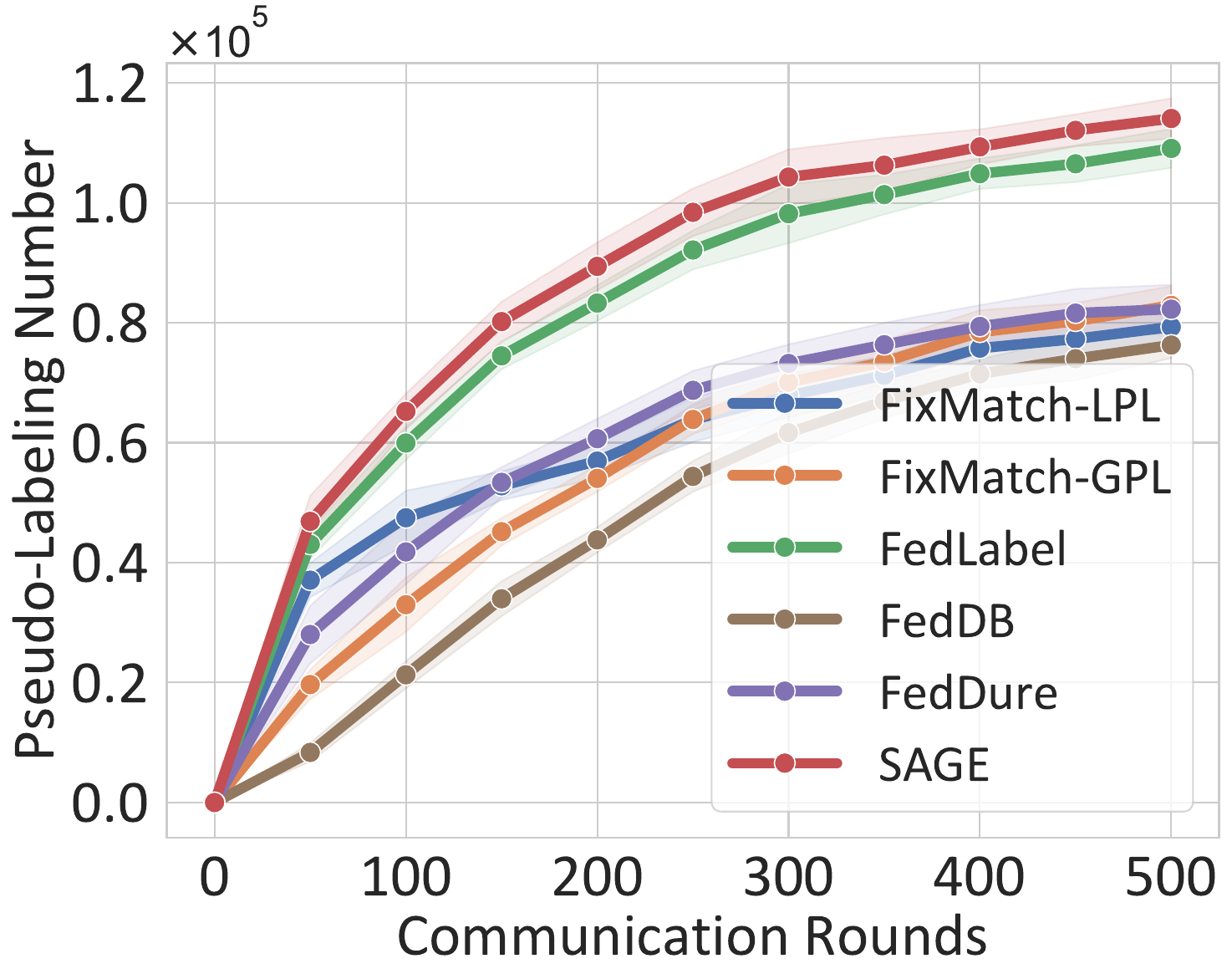}
        \caption{Comparison of pseudo-label counts on CIFAR-100.} 
        \label{fig:figure7}
    \end{minipage}
\end{figure*}

\subsection{Performance Comparison}
Tab.~\ref{table:experimental_10per} presents the accuracy of various methods across different datasets and under Non-IID settings with 10\% label. Under inter-client and intra-client imbalances, FixMatch-GPL outperforms FixMatch-LPL because the global model’s pseudo-label generation is unaffected by local data distributions. Most existing FSSL methods based on hard pseudo-labels provide limited performance improvements and, in some cases, perform worse than the vanilla FixMatch method on certain datasets. In contrast, SAGE significantly mitigates the impact of potentially incorrect pseudo-labels by integrating local and global model predictions, \textit{achieving the highest test accuracy across all datasets, with more substantial improvements as the heterogeneity increases.} On the SVHN dataset, SAGE even reaches the performance of fully labeled FedAvg-SL. We attribute this improvement to the enhanced generalization brought by data augmentation. Other labeling ratios are provided in Tab.~\ref{table:experimental_20per} in Appendix~\ref{subsection:labeling_ratio}, \textit{where SAGE also achieves the best performance.}

\renewcommand{\arraystretch}{1.0}

\begin{table}[t]

\centering
\caption{Comparison of convergence rates between SAGE and other baseline methods with $\alpha=1$.}
  \resizebox{1\linewidth}{!}{
\begin{tabular}{ccccccc}
\hline \hline
Acc.     & \multicolumn{2}{c}{30\%}              & \multicolumn{2}{c}{40\%}              & \multicolumn{2}{c}{50\%}              \\
Method   & Round$\downarrow$ & Speedup$\uparrow$ & Round$\downarrow$ & Speedup$\uparrow$ & Round$\downarrow$ & Speedup$\uparrow$ \\ \hline \hline
LPL      & 118               & $\times$1.00      & 267               & $\times$1.00      & 527               & $\times$1.00      \\
GPL      & 94                & $\times$1.26      & 183               & $\times$1.46      & 390               & $\times$1.35      \\
FedLabel & 91                & $\times$1.30      & 164               & $\times$1.63      & 341               & $\times$1.55      \\
FedDB    & 103               & $\times$1.15      & 237               & $\times$1.13      & 418               & $\times$1.26      \\
FedDure  & 95                & $\times$1.24      & 182               & $\times$1.47      & 450               & $\times$1.17      \\ 
\textbf{SAGE} &
  \cellcolor[HTML]{EFEFEF}\textbf{56} &
  \cellcolor[HTML]{EFEFEF}\textbf{$\times$2.11} &
  \cellcolor[HTML]{EFEFEF}\textbf{112} &
  \cellcolor[HTML]{EFEFEF}\textbf{$\times$2.38} &
  \cellcolor[HTML]{EFEFEF}\textbf{242} &
  \cellcolor[HTML]{EFEFEF}\textbf{$\times$2.18} \\ \hline \hline 
\end{tabular}
}
  \label{table:cover_α=1}
\end{table}

\begin{table}[t]
% \vspace{-0.1cm}
\centering
  \caption{Module ablation studies on CPG and CDSC.}
%  \vspace{-0.1cm}
  \resizebox{1\linewidth}{!}{
\begin{tabular}{cccccccc}
\hline \hline
\multirow{2}{*}{CPG} & \multirow{2}{*}{CDSC} & \multicolumn{3}{c}{CIFAR100}                     & \multicolumn{3}{c}{CINIC10}                      \\
                     &                       & $\alpha=0.1$   & $\alpha=0.5$   & $\alpha=1$     & $\alpha=0.1$   & $\alpha=0.5$   & $\alpha=1$     \\ \hline \hline
           &            & 49.32 & 49.67 & 49.55 & 68.02 & 70.67 & 72.69   \\
\checkmark &            & 52.25 & 53.85 & 53.50 & 72.19 & 73.14 & 73.91 \\
           & \checkmark & 52.43 & 53.17 & 53.48 & 72.83 & 73.22 & 74.10 \\
\checkmark           & \checkmark            & \textbf{54.18} & \textbf{55.82} & \textbf{56.06} & \textbf{74.59} & \textbf{75.74} & \textbf{76.68} \\ \hline \hline
\end{tabular}
}

  \label{table:module_ablation}
\end{table}

\subsection{Convergence Rate}
\label{subsection:Convergence Rate}
As shown in Fig.~\ref{fig:figure4} and Tab.~\ref{table:cover_α=1}, SAGE significantly speeds up the convergence rate and test accuracy on the CIFAR-100 dataset when $\alpha=1$ (Other heterogeneous scenarios are similar and are provided in Appendix~\ref{subsection:convergence_rate}). Compared to baseline and existing FSSL methods, \textit{SAGE achieves higher accuracy within fewer communication rounds.} Existing FSSL methods based on hard pseudo-label strategies amplify the impact of incorrect pseudo-labels, leading to greater divergence of local models under non-IID conditions. In contrast, SAGE dynamically corrects pseudo-labels using the global model, establishing stronger consensus between local and global models, thereby accelerating model convergence in the early stages of training.

\begin{table}[!t]

\centering
\caption{Performance gains brought by SAGE as a plugin to other baseline methods.}
  \resizebox{1\linewidth}{!}{
\begin{tabular}{ccccccc}
\hline \hline
 &
  \multicolumn{3}{c}{CIFAR-100} &
  \multicolumn{3}{c}{SVHN} \\
\multirow{-2}{*}{Methods} &
  $\alpha=0.1$ &
  $\alpha=0.5$ &
  $\alpha=1$ &
  $\alpha=0.1$ &
  $\alpha=0.5$ &
  $\alpha=1$ \\ \hline \hline
\multicolumn{1}{l}{\textbf{FixMatch}} &
  49.32 &
  49.67 &
  49.55 &
  90.46 &
  91.36 &
  91.25 \\
 &
  \cellcolor[HTML]{EFEFEF}54.18 &
  \cellcolor[HTML]{EFEFEF}55.82 &
  \cellcolor[HTML]{EFEFEF}56.06 &
  \cellcolor[HTML]{EFEFEF}93.85 &
  \cellcolor[HTML]{EFEFEF}94.27 &
  \cellcolor[HTML]{EFEFEF}94.65 \\
\multirow{-2}{*}{+SAGE} &
  {\color[HTML]{009901} ↑ 4.86} &
  {\color[HTML]{009901} ↑ 6.15} &
  {\color[HTML]{009901} ↑ 6.51} &
  {\color[HTML]{009901} ↑ 3.39} &
  {\color[HTML]{009901} ↑ 2.91} &
  {\color[HTML]{009901} ↑ 3.40} \\ \hline \hline
\multicolumn{1}{l}{\textbf{FlexMatch}} &
  49.91 &
  51.39 &
  51.79 &
  52.58 &
  55.59 &
  60.50 \\
 &
  \cellcolor[HTML]{EFEFEF}49.84 &
  \cellcolor[HTML]{EFEFEF}51.41 &
  \cellcolor[HTML]{EFEFEF}52.06 &
  \cellcolor[HTML]{EFEFEF}93.36 &
  \cellcolor[HTML]{EFEFEF}94.26 &
  \cellcolor[HTML]{EFEFEF}93.86 \\
\multirow{-2}{*}{+SAGE} &
  {\color[HTML]{CB0000} ↓ 0.07} &
  {\color[HTML]{009901} ↑ 0.02} &
  {\color[HTML]{009901} ↑ 0.27} &
  {\color[HTML]{009901} ↑ 40.78} &
  {\color[HTML]{009901} ↑ 38.67} &
  {\color[HTML]{009901} ↑ 33.36} \\ \hline \hline
\multicolumn{1}{l}{\textbf{FedDure}} &
  48.27 &
  51.09 &
  50.79 &
  92.87 &
  93.49 &
  94.19 \\
 &
  \cellcolor[HTML]{EFEFEF}54.13 &
  \cellcolor[HTML]{EFEFEF}56.23 &
  \cellcolor[HTML]{EFEFEF}55.84 &
  \cellcolor[HTML]{EFEFEF}93.96 &
  \cellcolor[HTML]{EFEFEF}94.11 &
  \cellcolor[HTML]{EFEFEF}94.31 \\
\multirow{-2}{*}{+SAGE} &
  {\color[HTML]{009901} ↑ 5.86} &
  {\color[HTML]{009901} ↑ 5.14} &
  {\color[HTML]{009901} ↑ 5.05} &
  {\color[HTML]{009901} ↑ 1.09} &
  {\color[HTML]{009901} ↑ 0.62} &
  {\color[HTML]{009901} ↑ 0.12} \\ \hline \hline
\multicolumn{1}{l}{\textbf{FedDB}} &
  48.43 &
  50.11 &
  51.55 &
  92.56 &
  93.00 &
  94.14 \\
 &
  \cellcolor[HTML]{EFEFEF}48.33 &
  \cellcolor[HTML]{EFEFEF}50.27 &
  \cellcolor[HTML]{EFEFEF}51.84 &
  \cellcolor[HTML]{EFEFEF}92.51 &
  \cellcolor[HTML]{EFEFEF}93.16 &
  \cellcolor[HTML]{EFEFEF}93.42 \\
\multirow{-2}{*}{+SAGE} &
  {\color[HTML]{CB0000} ↓ 0.10} &
  {\color[HTML]{009901} ↑ 0.16} &
  {\color[HTML]{009901} ↑ 0.29} &
  {\color[HTML]{CB0000} ↓ 0.05} &
  {\color[HTML]{009901} ↑ 0.16} &
  {\color[HTML]{009901} ↑ 0.28} \\ \hline \hline
\end{tabular}
}
  \label{table:table_plugin}
\end{table}

\subsection{SAGE as a Plug-in Approach}
The CPL and CDSC components of SAGE function as pseudo-labeling mechanisms agnostic to local semi-supervised training specifics, allowing integration as plugins into hard pseudo-labeling-based SSL and FSSL methods. As shown in Tab.~\ref{table:table_plugin}, \textit{SAGE improves the performance of existing methods.} This is especially beneficial for FlexMatch, which, due to its strategy of dynamically adjusting class thresholds, is prone to overfitting under class imbalance, a problem exacerbated in non-IID settings. SAGE mitigates this issue by incorporating global information into the pseudo-labeling strategy, resulting in significant performance improvements for FlexMatch on the SVHN dataset.

\subsection{Ablation Study}
\label{sec:ablation_study}
In this section, we conduct an in-depth ablation study to demonstrate the contributions of CPG and CDSC within SAGE. More ablation studies on hyperparameter tuning and experiments under different heterogeneity are provided in Appendix.~\ref{section:Additional_ablation_Study}.

\paragraph{Effectiveness of Components.}
We first validated the contributions of CPG and CDSC through ablation experiments. FedAvg+FixMatch-LPL, the vanilla combination of FedAvg with FixMatch, served as the baseline method. Experiments were conducted on client data with different levels of data heterogeneity $\alpha = \{0.1, 0.5, 1\}$ to assess component effectiveness. As shown in Tab.~\ref{table:module_ablation}, \textit{each component consistently enhances model performance under different levels of heterogeneity.} With both CPG and CDSC included, SAGE achieves the best performance gain.

\begin{figure}[!t]
  \centering
  \begin{subfigure}{0.49\linewidth}
    \centering
    \includegraphics[width=\linewidth]{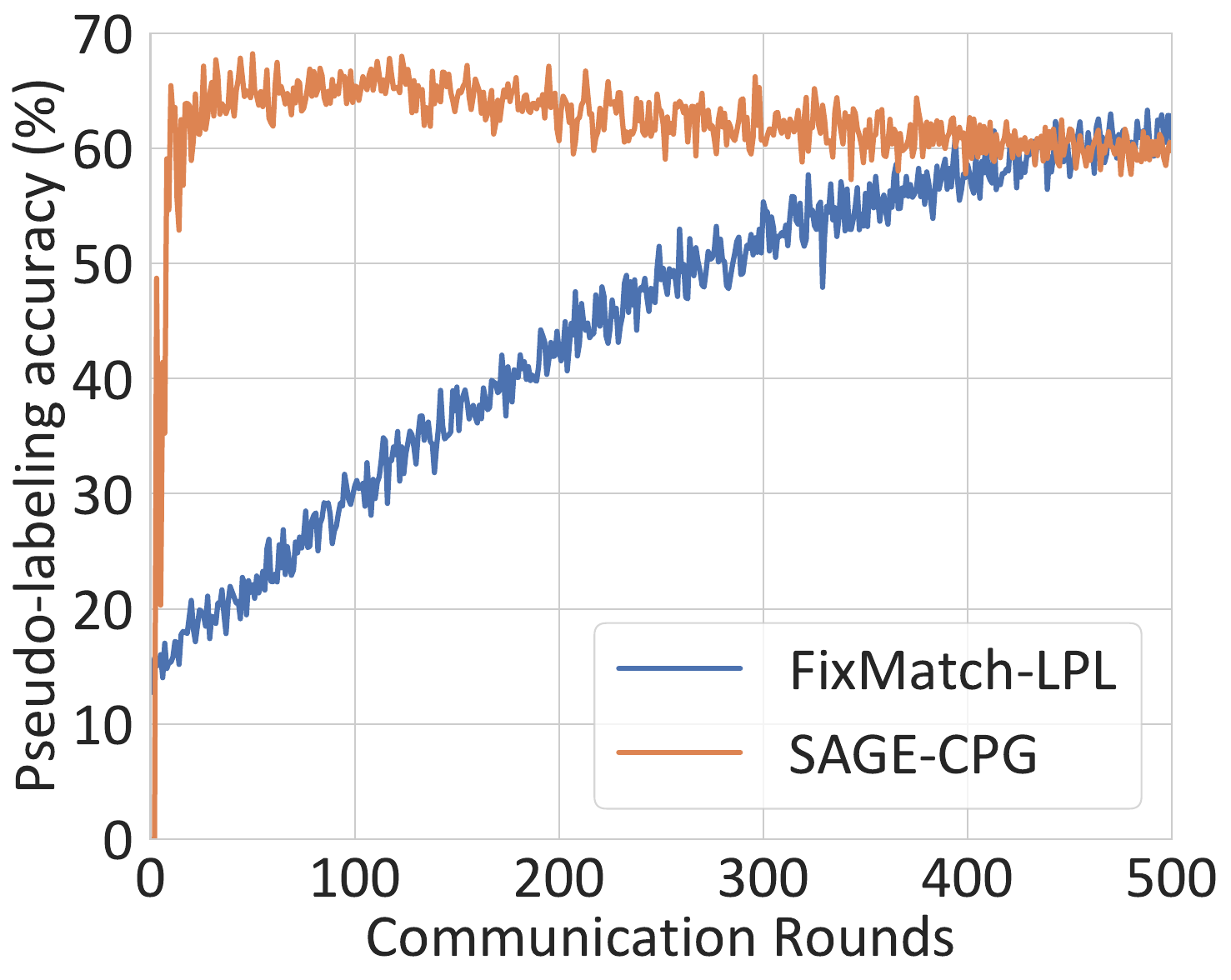}
    \caption{Comparison of pseudo-labeling Acc. between CPG and baseline.}
    \label{fig:figure6a}
  \end{subfigure}
  \hfill
  \begin{subfigure}{0.49\linewidth}
    \centering
    \includegraphics[width=\linewidth]{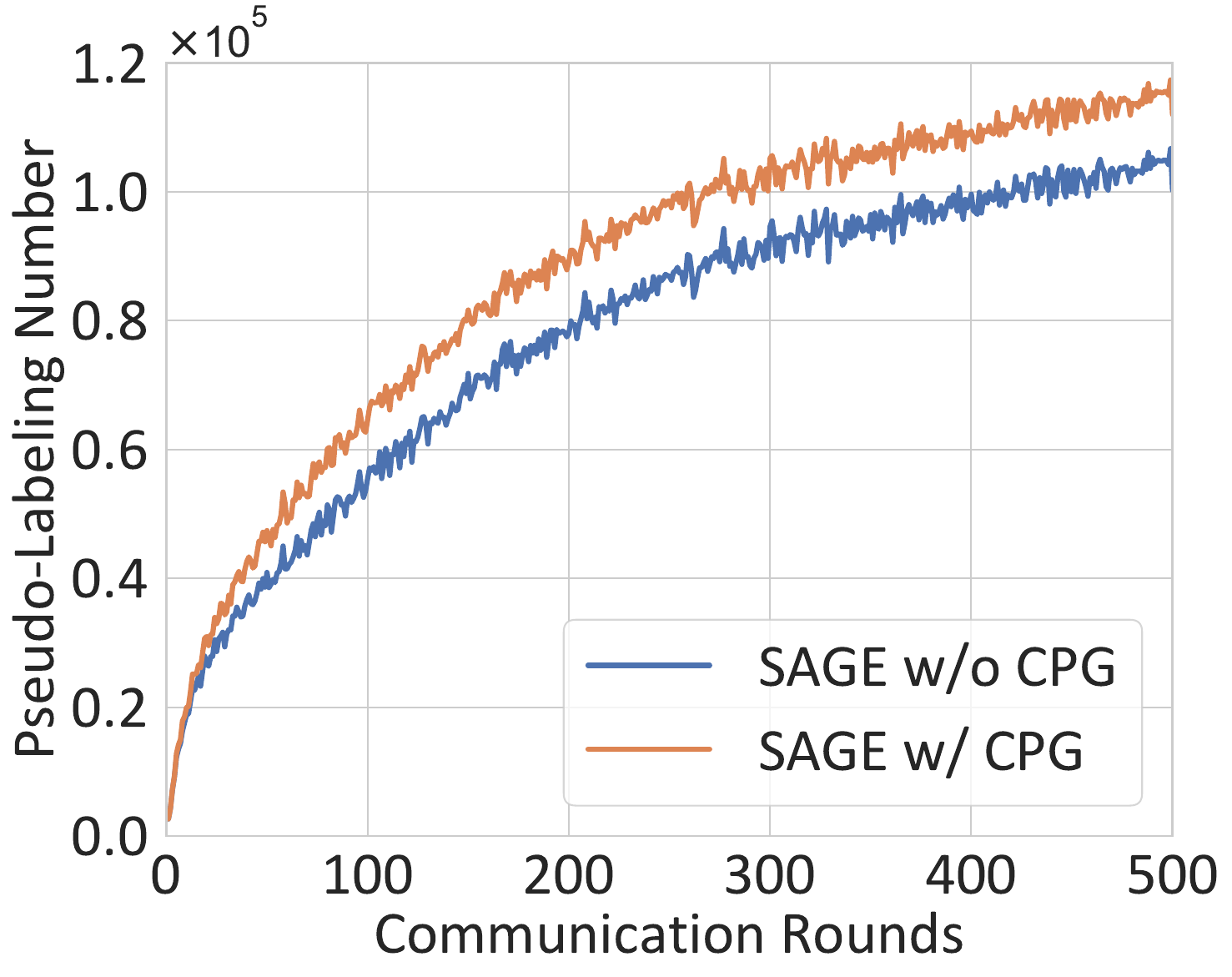}
    \caption{The increase in pseudo-labels generated by CPG in SAGE.}
    \label{fig:figure6b}
  \end{subfigure}
  
  \caption{In-depth ablation of CPG on CIFAR-100. CPG significantly increases the utilization of unlabeled data of SAGE while ensuring pseudo-labeling accuracy.}
  \label{fig:figure6}
\end{figure}

\paragraph{Pseudo-label Gains from CPG.} 
We monitor the number of pseudo-labels generated by SAGE and the baselines throughout training. As shown in Fig.~\ref{fig:figure7}, with the enhancement provided by CPG, \textit{SAGE consistently maintains a lead in pseudo-label count}, a key factor in SAGE’s performance improvement. We conducted a further analysis of the performance gains from CPG. As shown in Fig.~\ref{fig:figure6}, compared to a single local pseudo-labeling strategy, CPG generates high-accuracy pseudo-labels early in training. With the assistance of the global model, \textit{CPG effectively compensates for the scarcity of pseudo-labels in local minority classes, further enhancing the utilization of unlabeled data}.

\paragraph{Dynamic Correction Coefficient $\lambda$.}

In CDSC, the correction coefficient $\lambda(x)$ quantifies prediction discrepancy between local and global models, balancing the confident predictions of the local model with the conservative predictions of the global model. To evaluate its effectiveness, we compared the dynamic coefficient against fixed values of $\lambda$ (ranging from 0 to 1). When $\lambda=0$, the method reduces to FixMatch-LPL, relying only on local pseudo-labels; when $\lambda=1$, it relies solely on global pseudo-labels, as in FixMatch-GPL. Experimental results in Fig.~\ref{fig:figure5} demonstrate that \textit{regardless of the fixed value of $\lambda$, the model’s performance surpasses both FixMatch-LPL and FixMatch-GPL, but does not achieve the effectiveness of the dynamic $\lambda$. }This finding suggests that assigning a greater global weight to samples with larger confidence discrepancies can more effectively mitigate the impact of potentially incorrect pseudo-labels and thus improve model performance. More ablation studies on $\lambda$ are provided in Appendix~\ref{sec:additional_ablatioin_lambda}.

\begin{figure}[!t]
  \centering
  \begin{subfigure}{0.49\linewidth}
    \centering
    \includegraphics[width=\linewidth]{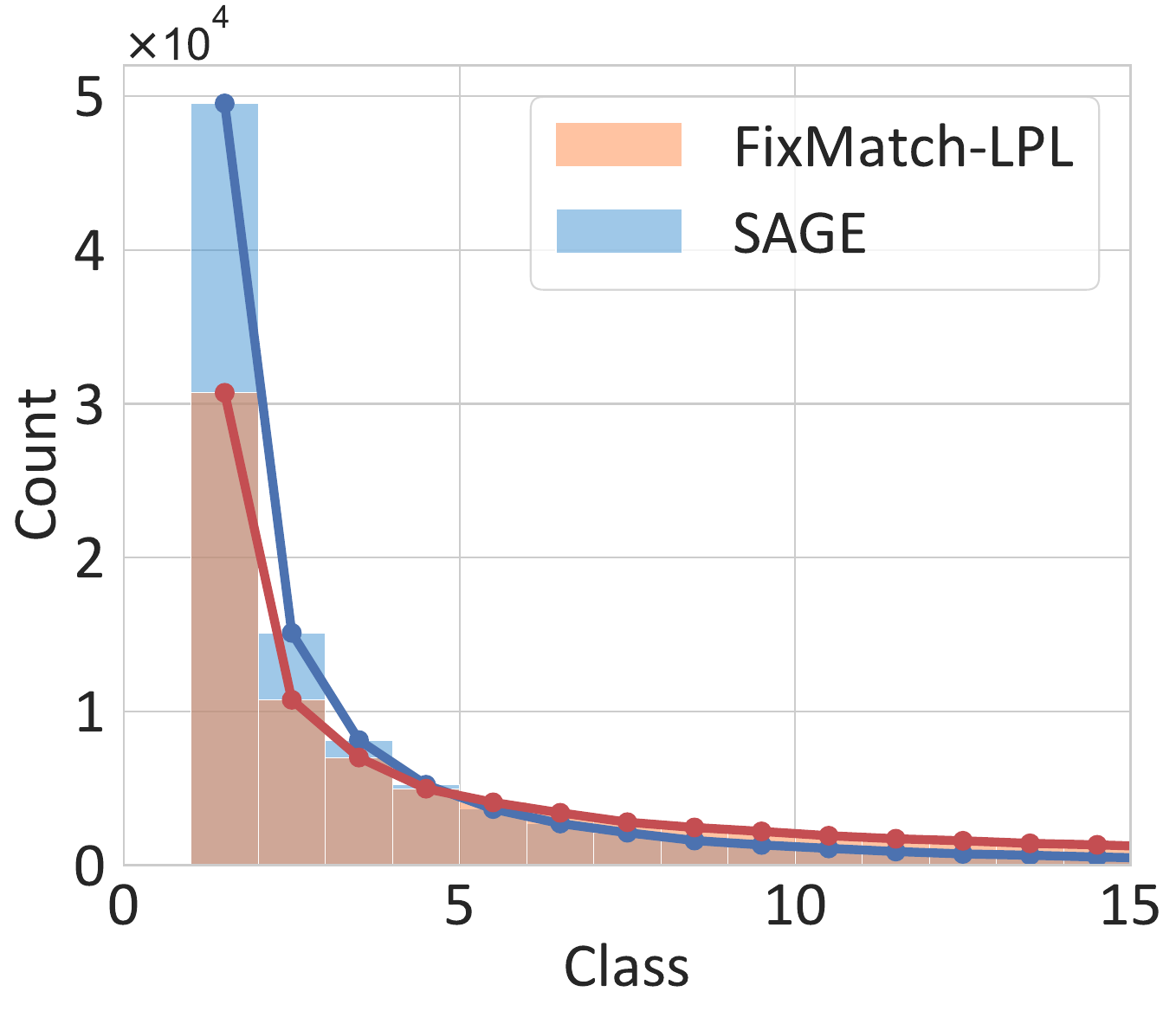}
    % The ranking of local model pseudo-label predictions within the global model predictions.
    \caption{Ranking in global predictions.}
    \label{fig:figure8a}
  \end{subfigure}
  \hfill
  \begin{subfigure}{0.49\linewidth}
    \centering
    \includegraphics[width=\linewidth]{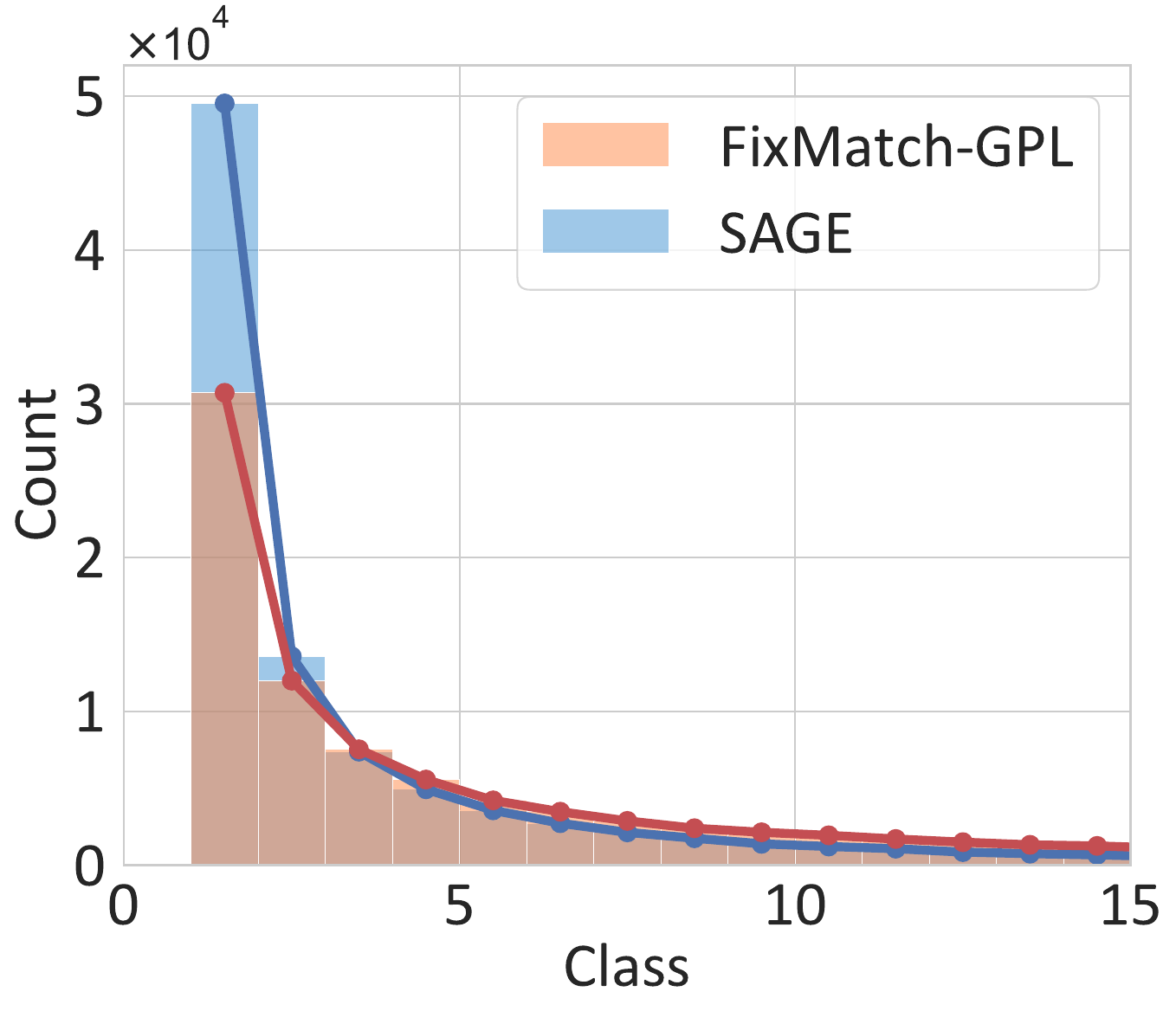}
    % The ranking of global model pseudo-label predictions within the local model predictions.
    \caption{Ranking in local predictions.}
    \label{fig:figure8b}
  \end{subfigure}
  
  \caption{Consensus ablation between local and global models. (a) displays the ranking statistics of the local model's pseudo-labels within the global model's class predictions, while (b) displays the ranking statistics of the global model's pseudo-labels within the local model's class predictions.}
  \label{fig:figure8}
\end{figure}

\paragraph{CDSC Enhances Consensus Between Global and Local.} As stated in Remark~\hyperlink{remark1}{1}, existing FSSL methods based on hard pseudo-labels cause local models to fit local biased distributions more aggressively, amplifying the discrepancy between global and local models. Fig.~\ref{fig:figure8} presents a histogram of predicted class rankings, demonstrating the improvement in predictive consensus achieved by SAGE. Taking Fig.~\ref{fig:figure8}(a) as an example, after applying SAGE the pseudo-label predictions of the local model tend to rank higher within the global model's class predictions. Similarly, Fig.~\ref{fig:figure8}(b) exhibits that the predictions of the global model exhibit the same trend. This indicates that \textit{SAGE effectively reduces prediction discrepancies between local and global models, thereby enhancing their consensus and accelerating the model convergence.}

\section{Conclusion}
In this study, it was initially observed that increasing heterogeneity leads to pseudo-label mismatches in FSSL, which subsequently affect model performance and convergence. Another intriguing phenomenon was discovered: as heterogeneity increases, the confidence discrepancy between the local and global models expands.We analyzed the underlying rationale and, based on this observation, proposed a new approach called SAGE.SAGE leverages confidence discrepancies for flexible pseudo-label correction, enhancing the utilization of unlabeled data, mitigating the adverse effects of incorrect pseudo-labels, and strengthening the consensus between local and global models. In future work, we aim is to extend the applicability of SAGE to ensure robust performance across different FSSL scenarios, including Label-at-Partial-Client and Label-at-Server settings.Client and Label-at-Server settings.

\section*{Acknowledgements}
\textls[-5]{This study was supported in part by the National Natural Science Foundation of China under Grants 62376233, 62431004, 62476063, U21A20514, 62336003 and 12371510; in part by the Natural Science Foundation of Fujian Province under Grant 2024J09001; and in part by Xiaomi Young Talents Program.}

{
    \small
    \bibliographystyle{ieeenat_fullname}
    \normalem
    \bibliography{main}
}

% =====================appendix=====================
\clearpage
\setcounter{page}{1}
\maketitlesupplementary
\appendix

\renewcommand{\arraystretch}{1.2} 

\section{Pseudo-Code of SAGE}
\label{sec:workflow}
The pseudo-code of SAGE is shown in Algorithm~\ref{algorithm:sage}.

\normalem
\begin{algorithm}[h]
\setstretch{1.0}
    \caption{\textbf{S}emi-supervised \textbf{A}ggregation for \textbf{G}lobally-Enhanced \textbf{E}nsemble (SAGE)}
    \label{algorithm:sage}
    
    \SetAlgoLined
    
    \KwIn{\justifying Set of clients $\mathcal{C}$; number of online clients in each round $M$; number of communication rounds $T$; number of local training epochs $E$; weak augmentation $\alpha(\cdot)$; strong augmentation $\mathcal{A}(\cdot)$; confidence threshold $\tau$; learning rate $\gamma$; unsupervised loss weight $\mu_u$; dynamic correction coefficient $\lambda(\cdot)$; sensitivity hyperparameter $\kappa$}
    
    \BlankLine
    \textbf{ServerExecutes:}\\
    Randomly initialize global model parameters $\theta_g$\;
    \For{$t = 0$ \KwTo $T-1$}{
        Randomly select online clients $\mathcal{C}_M \subseteq \mathcal{C}$\;
        \ForEach{client $C_m \in \mathcal{C}_M$ \textbf{in parallel}}{
            $\theta_{l,m} \leftarrow$ \textbf{ClientUpdate}($\theta_g$)
        }
        $|D| = \sum_{C_m \in \mathcal{C_M}}(|\mathcal{D}^s_m|+|\mathcal{D}^u_m|)$\;
        $\theta_g \leftarrow  \frac{1}{|D|}\cdot \sum_{C_m \in \mathcal{C_M}} ((|\mathcal{D}^s_m| + |\mathcal{D}^u_m|)\cdot \theta_{l,m})$\;
    }
    \Return{$\theta_g^T$}

    \BlankLine

    \textbf{ClientUpdate}($\theta_g$):\\
    $\theta_l \leftarrow \theta_g$\;
    \For{$e = 0$ \KwTo $E-1$}{
        \ForEach{$(\mathbf{x}, \mathbf{y}) \in \mathcal{D}^s, \mathbf{u} \in \mathcal{D}^u$}{
            $\mathcal{L}_s \leftarrow \mathcal{L}_{CE} (p_l(y | \mathbf{x}, \mathbf{y}))$\;
            $p_l \leftarrow f_l(\alpha (\mathbf{u}))$\;
            $p_g \leftarrow f_g(\alpha (\mathbf{u}))$\;
            Calculate $\hat{y}$ by CPG in Eq.~\eqref{eq:collaborative_pseudo_generation};\par
            \If{$\max(p_l) \geq \tau$}{
                $\Delta C = |\max(p_l) - \max(p_g)|$\;
                $\lambda \leftarrow \exp(-\kappa \cdot \Delta C)$\;
                $\delta_l \leftarrow \text{one-hot}(\arg \max(p_l))$\;
                $\delta_g \leftarrow \text{one-hot}(\arg \max(p_g))$\;
                Calculate $\hat{y}$ by CDSC in Eq.~\eqref{eq:cdsc_pseudo_label}\;
            }
            $L_u \leftarrow \text{KL}(p_{l}(\mathcal{A}(\mathbf{u})) \parallel \hat{y}(\mathbf{u}))$\;
            $\theta_l \leftarrow \theta_l - \gamma \nabla_{\theta}(L_s + \mu_u \cdot L_u)$\;
        }
    }
    \Return{$\theta_l$, $\mathcal{D}^s$, $\mathcal{D}^u$}
\end{algorithm}
\ULforem

In the local training process of SAGE, standard supervised training is initially performed on labeled data (line 16) to compute $L_s$. Next, CPG assigns initial pseudo-labels $\hat{y}$ using Eq.~\eqref{eq:collaborative_pseudo_generation} (lines 16 to 19), thereby enhancing the utilization of unlabeled data. Subsequently, the confidence discrepancy $\Delta C$ between the local and global models is calculated, and the pseudo-labels are dynamically refined by computing the correction coefficient $\lambda$ (lines 20 to 25) using CDSC. Finally, the KL divergence between the corrected pseudo-labels and the strongly augmented predictions of the local model is calculated as the unsupervised loss $L_u$. Upon completing local training, clients upload the updated local models and dataset sizes to the server for standard federated aggregation (lines 4 to 9).

\section{Additional Analysis of Preliminary Study}
In Section~\ref{section:preliminary_study}, we identified an intriguing phenomenon: as data heterogeneity increases, the confidence discrepancy between local and global models progressively grows. The predictions of the local model become more aggressive, whereas those of the global model grow increasingly conservative, as described in Observation~\hyperlink{obs1}{1} and~\hyperlink{obs2}{2}. In this section, we perform a more comprehensive observation and analysis of this phenomenon. First, we provide additional observations in Appendix~\ref{sec:additional_exploratory_experiments}. Next, in Appendix~\ref{sec:analysis}, we derive the underlying causes of this phenomenon and present a analytical process centered on Remark~\hyperlink{remark1}{1} and~\hyperlink{remark2}{2}. Finally, in Appendix~\ref{sec:experimental_support}, we design experiments to validate our analytical conclusions.

\begin{figure}[t]
  \centering

  \begin{subfigure}{0.49\linewidth}
    \centering
    \includegraphics[width=\linewidth]{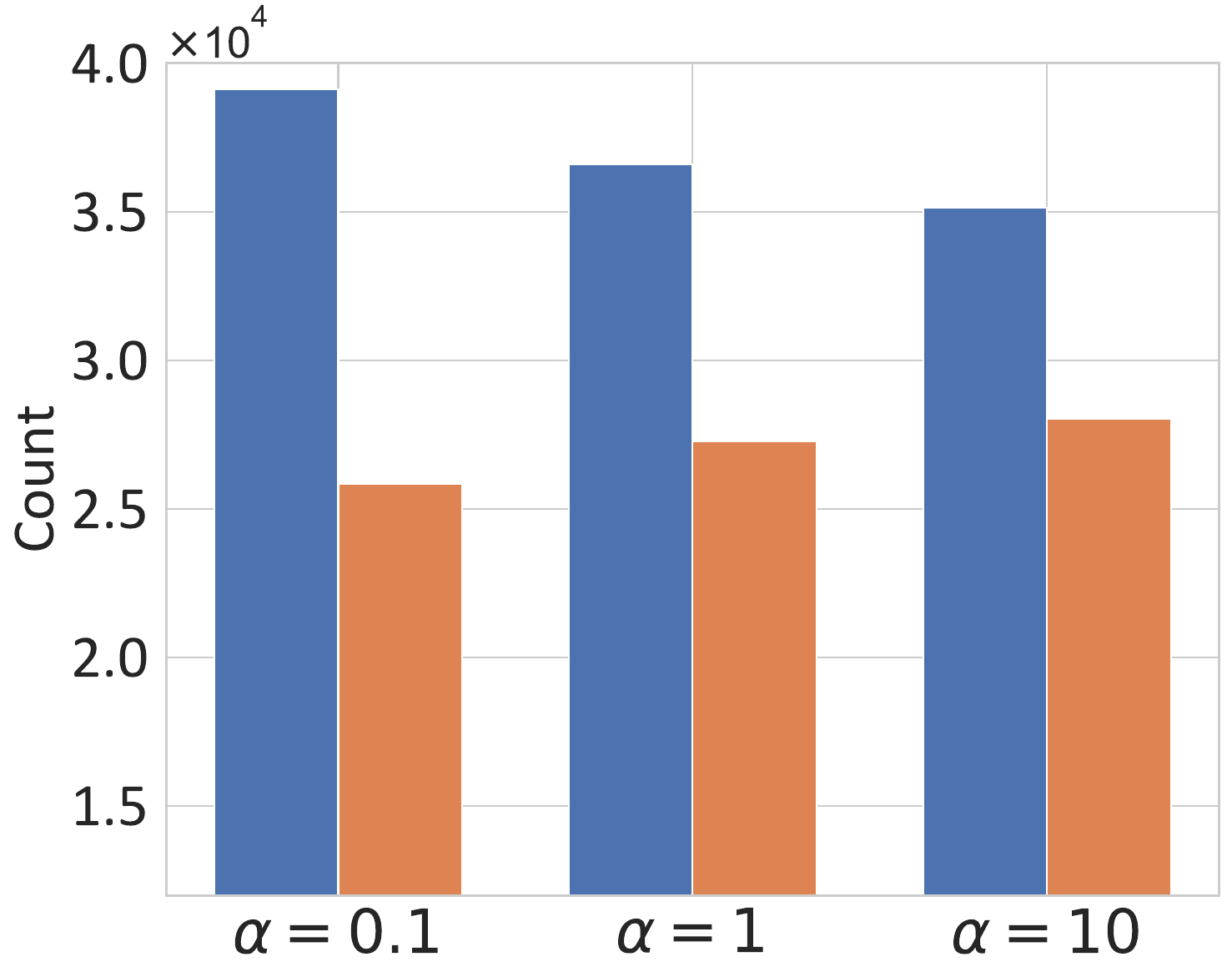}
    \caption{Confidence $>0.95$.}
    \label{fig:fig_a}
  \end{subfigure}
  \hfill
  \begin{subfigure}{0.49\linewidth}
    \centering
    \includegraphics[width=\linewidth]{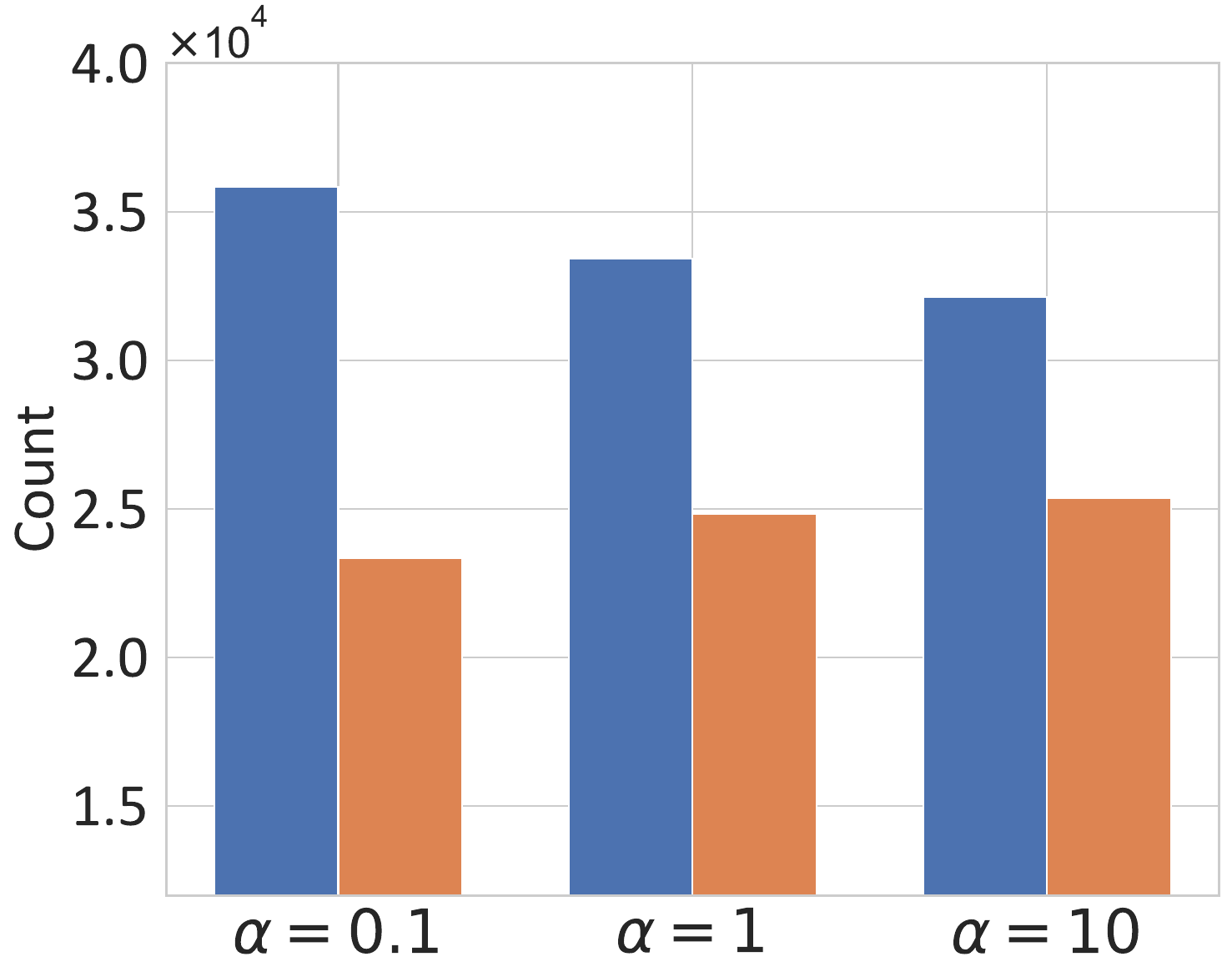}
    \caption{Confidence $>0.96$.}
    \label{fig:fig_b}
  \end{subfigure}

  \begin{subfigure}{0.49\linewidth}
    \centering
    \includegraphics[width=\linewidth]{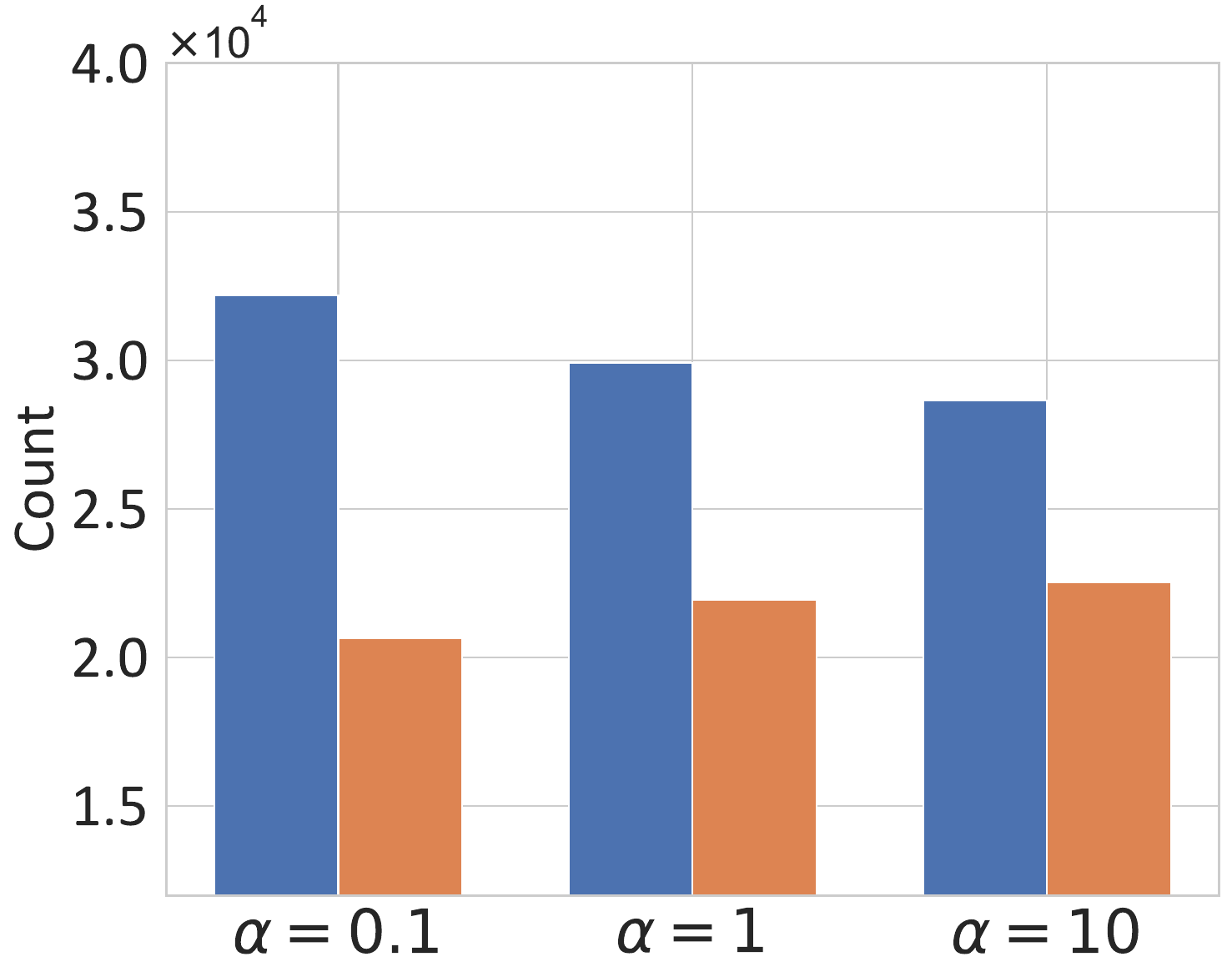}
    \caption{Confidence $>0.97$.}
    \label{fig:fig_c}
  \end{subfigure}
  \hfill
  \begin{subfigure}{0.49\linewidth}
    \centering
    \includegraphics[width=\linewidth]{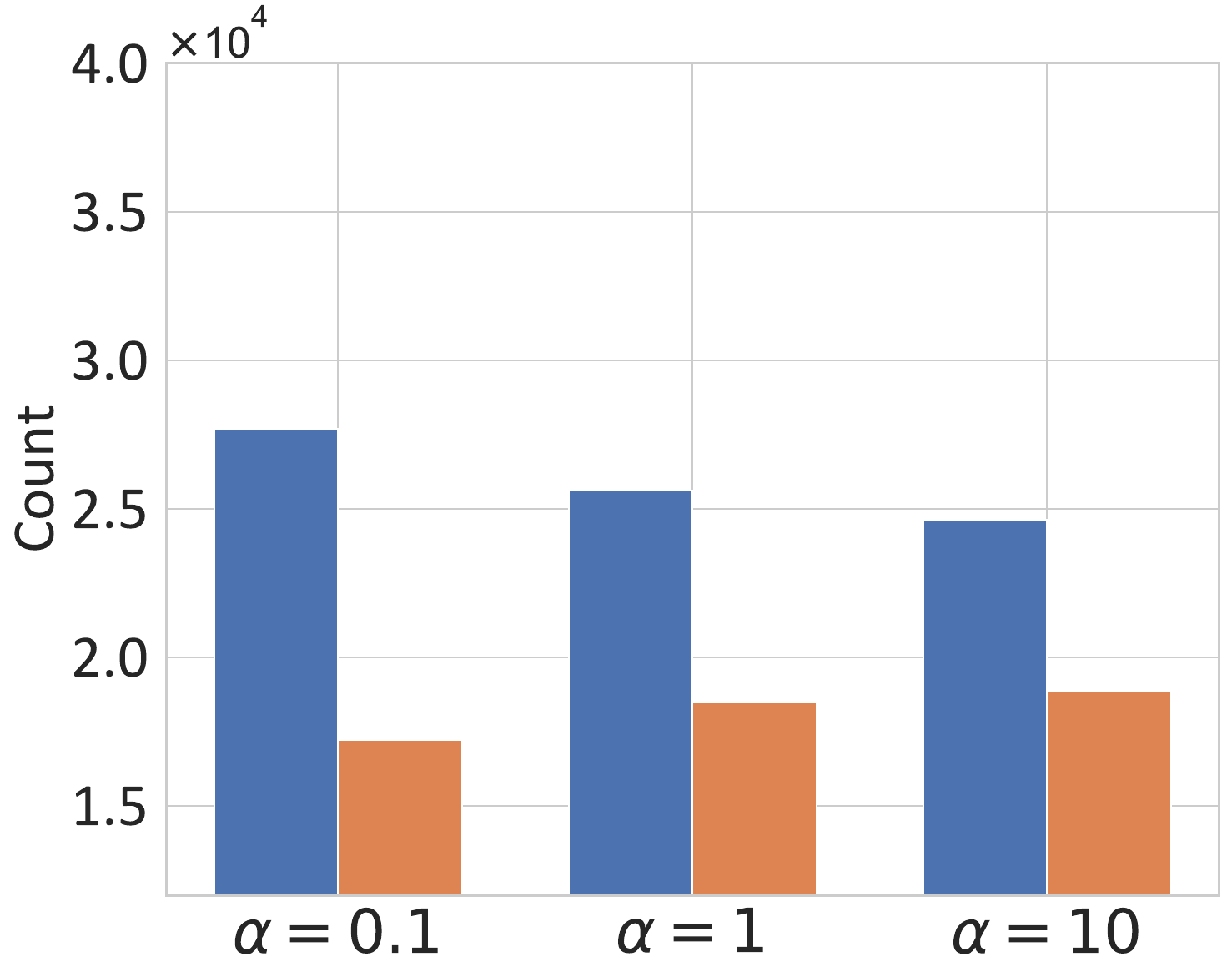}
    \caption{Confidence $>0.98$.}
    \label{fig:fig_d}
  \end{subfigure}

  \begin{subfigure}{\linewidth}
    \centering
    \includegraphics[width=\linewidth]{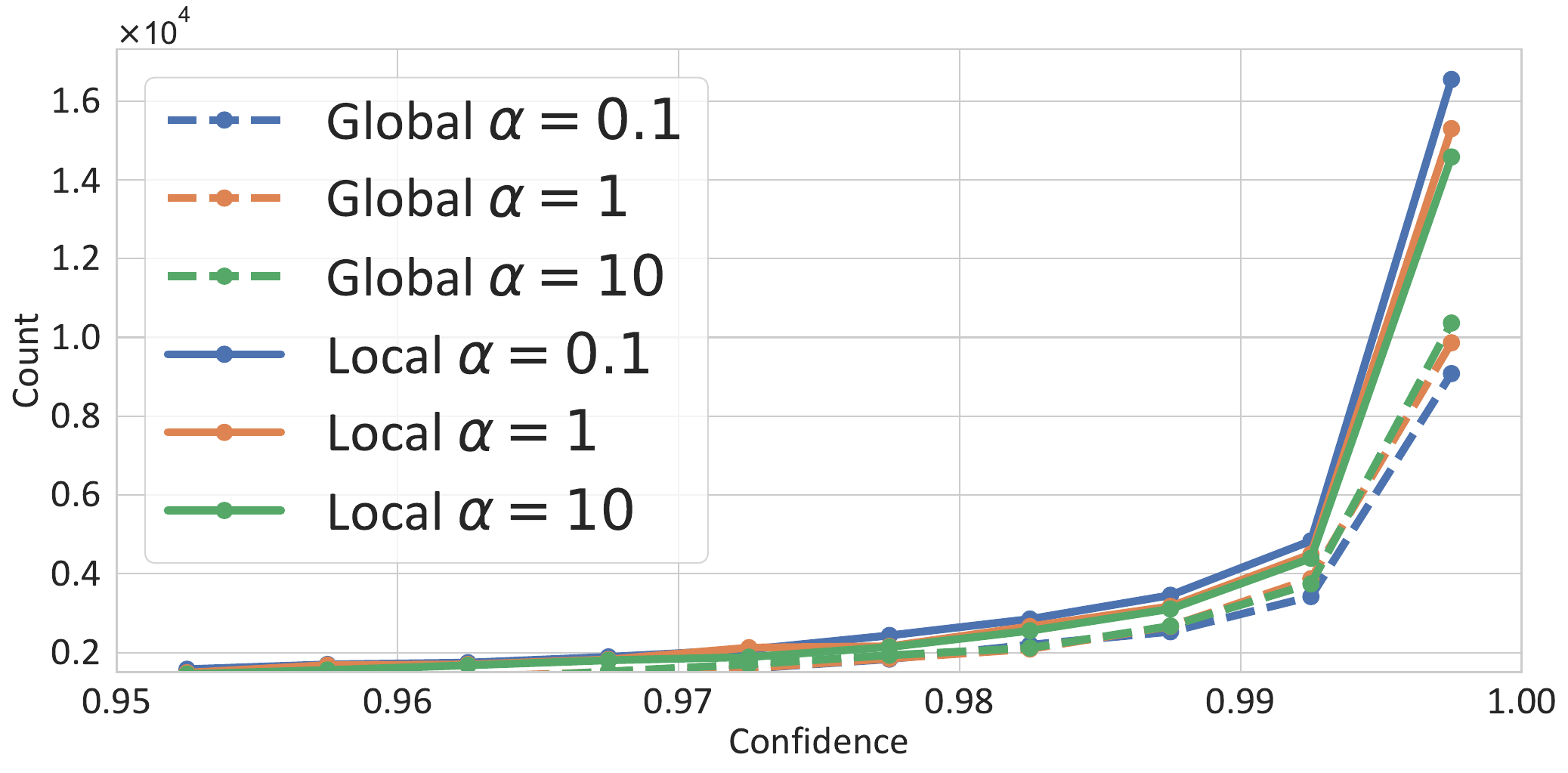}
    \caption{Line Chart of Confidence Distribution.}
    \label{fig:fig_e}
  \end{subfigure}

  \caption{Pseudo-label distribution of local and global models at different confidence distribution thresholds. Each subfigure represents a different threshold level, and the line chart shows the overall confidence distribution.}
  \label{fig:figure_additional_exploratory_experiments_a}
\end{figure}

\subsection{Additional Exploratory Experiments}
\label{sec:additional_exploratory_experiments}
To more comprehensively illustrate Observation~\hyperlink{obs1}{1} and \hyperlink{obs2}{2}, we follow the experimental setup of Fig.~\ref{fig:preliminary_study}(a) and adjust the threshold values for displaying confidence distributions. As shown in Fig.~\ref{fig:figure_additional_exploratory_experiments_a}, we observe similar patterns as in Fig.~\ref{fig:preliminary_study}(a) of the main text: as data heterogeneity increases, the confidence of the local model tends to fall into high-confidence regions, while the global model shows the opposite trend.

Additionally, to expand on the comparison of pseudo-label counts between local and global models in Fig.~\ref{fig:preliminary_study}(b), we conducted further experiments across different heterogeneity settings. As shown in Fig.~\ref{fig:figure_additional_exploratory_experiments_b}, at varying levels of heterogeneity, the local model consistently maintains a high utilization rate of unlabeled data in the early training stages.

\subsection{Analysis of Local-Global Discrepancies}
\label{sec:analysis}

In Section~\ref{section:preliminary_study}, we observed that as heterogeneity intensifies, the pseudo-labeling tendencies of the local and global models change in markedly different ways. These specific phenomena are detailed in Observations~\hyperlink{obs1}{1} and \hyperlink{obs}{2}. In this section, we analyze the underlying reasons.

\begin{figure}[t]
  \centering
  \begin{subfigure}{0.49\linewidth}
    \centering
    \includegraphics[width=\linewidth]{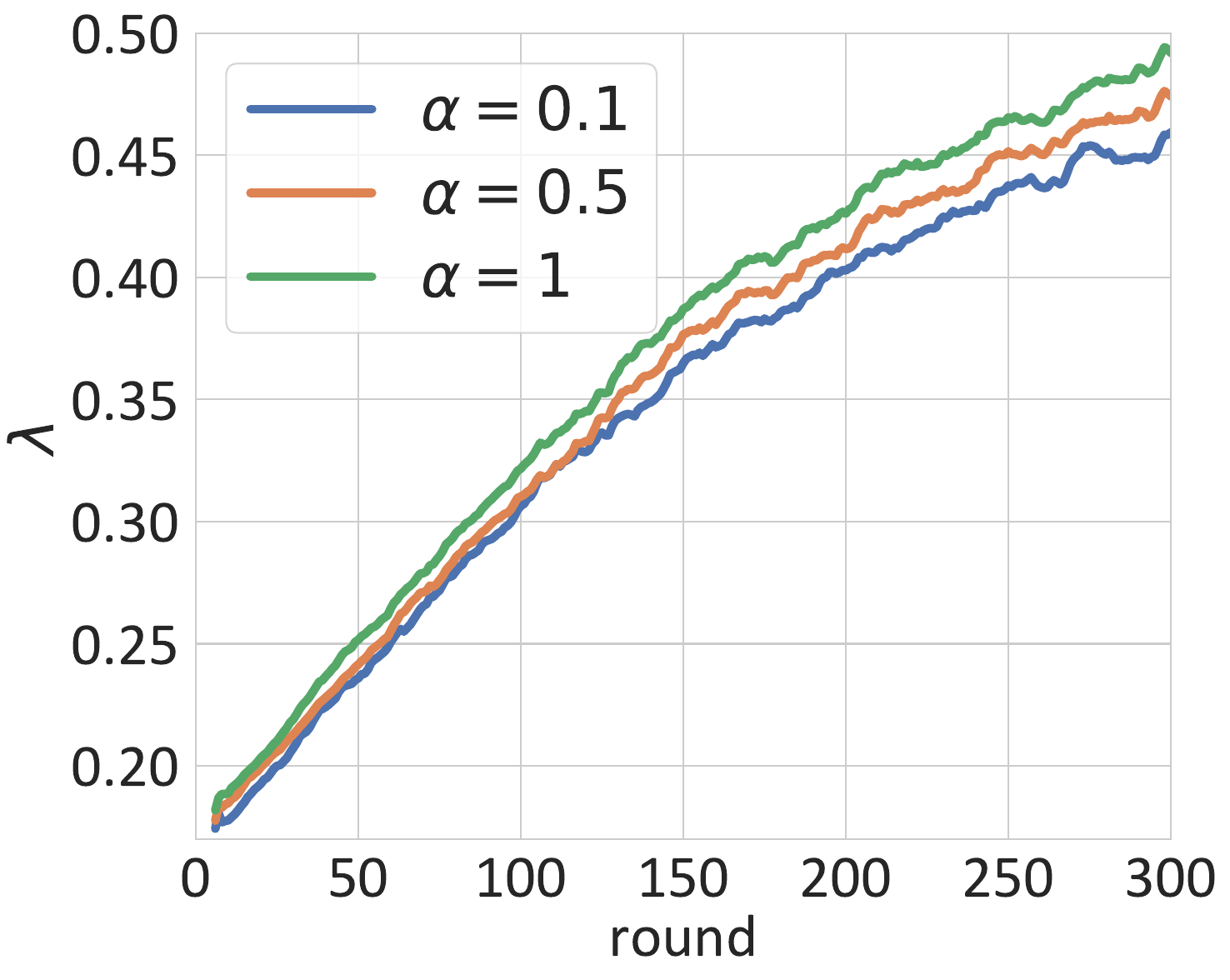}
    \caption{$\lambda$ under different data heterogeneity.}
    \label{fig:add_figure_lambda_1}
  \end{subfigure}
  \hfill
  \begin{subfigure}{0.49\linewidth}
    \centering
    \includegraphics[width=\linewidth]{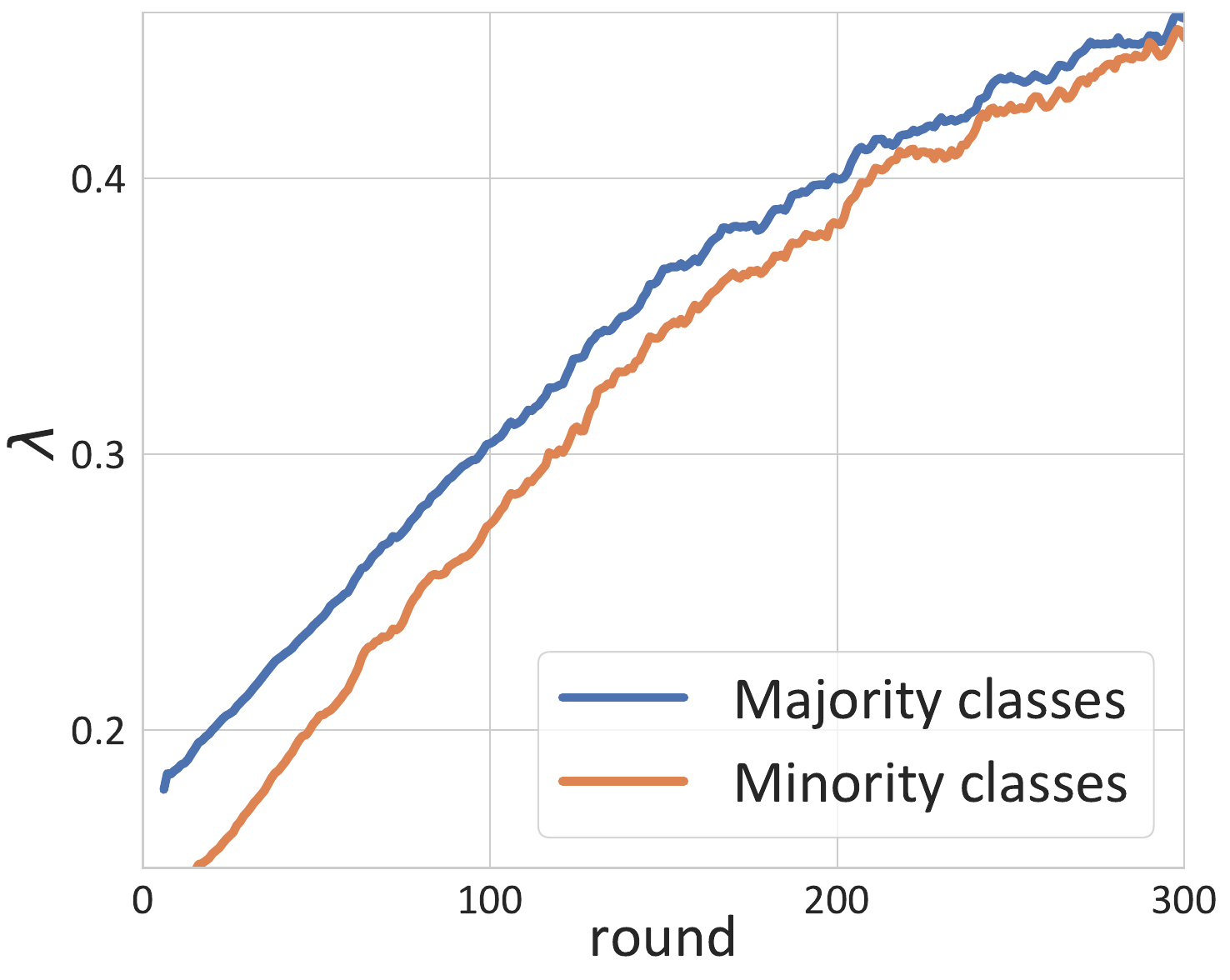}
    \caption{$\lambda$ under different data distribution.}
    \label{fig:add_figure_lambda_2}
  \end{subfigure}

    \caption{Ablation of $\lambda$ on CIFAR-100.}
  \label{fig:add_figure_lambda}
\end{figure}

\begin{figure}[t]
  \centering
  \begin{subfigure}{0.49\linewidth}
    \centering
    \includegraphics[width=\linewidth]{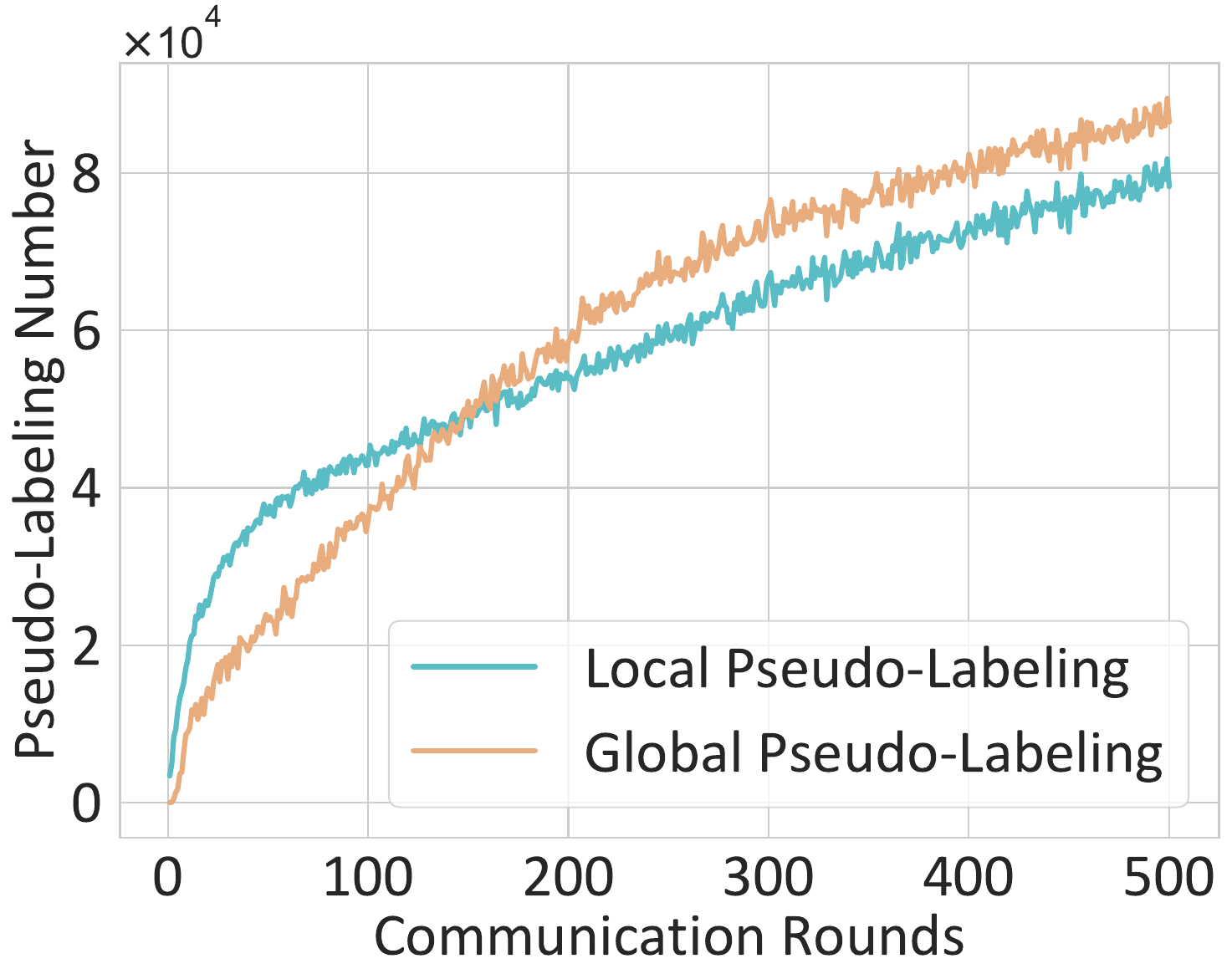}

    \caption{$\alpha=1.$}

  \end{subfigure}
  \hfill
  \begin{subfigure}{0.49\linewidth}
    \centering
    \includegraphics[width=\linewidth]{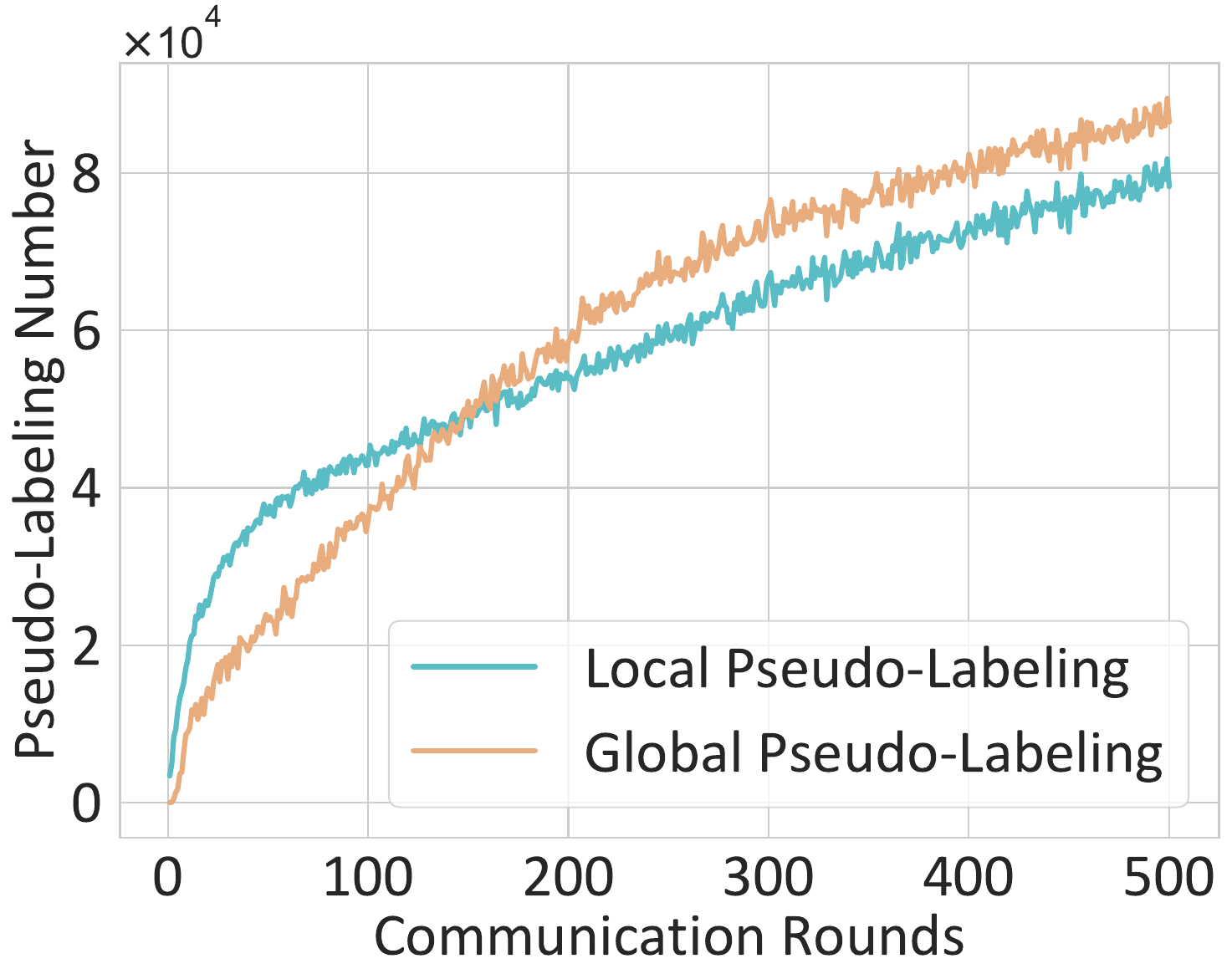}

    \caption{$\alpha=10.$}
  \end{subfigure}
  
  \caption{The number of pseudo labels for local and global models under the additional heterogeneity setting.}
  \label{fig:figure_additional_exploratory_experiments_b}
\end{figure}

\begin{figure*}[!t]
  \centering
  % first subfigure
  \begin{subfigure}{0.49\linewidth}
    \centering
    \includegraphics[width=\linewidth]{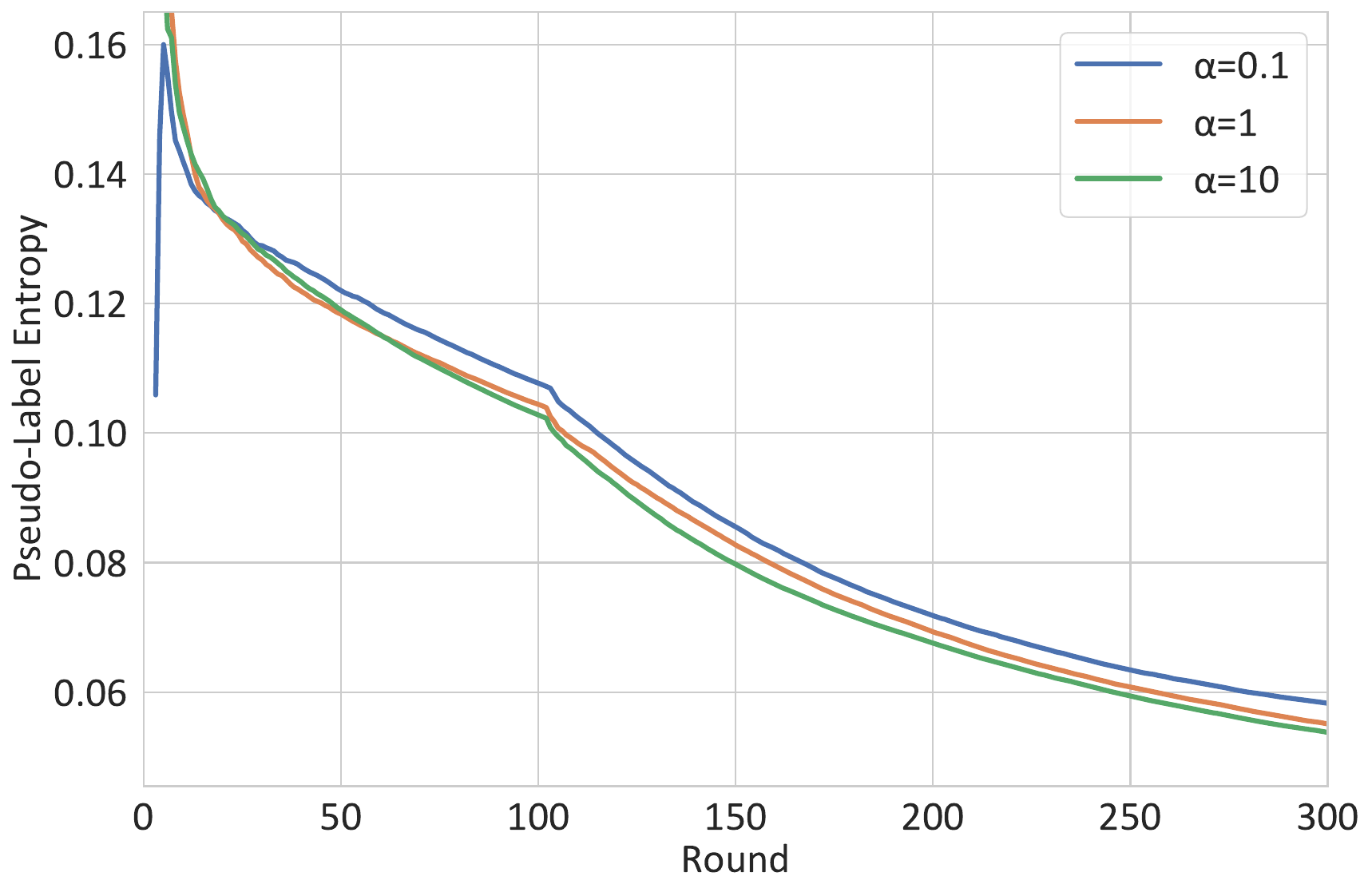}
    \caption{Pseudo-label entropy of the global model under different heterogeneity.}
  \end{subfigure}
    \hfill
  % second subfigure
    \begin{subfigure}{0.49\linewidth}
    \centering
    \includegraphics[width=\linewidth]{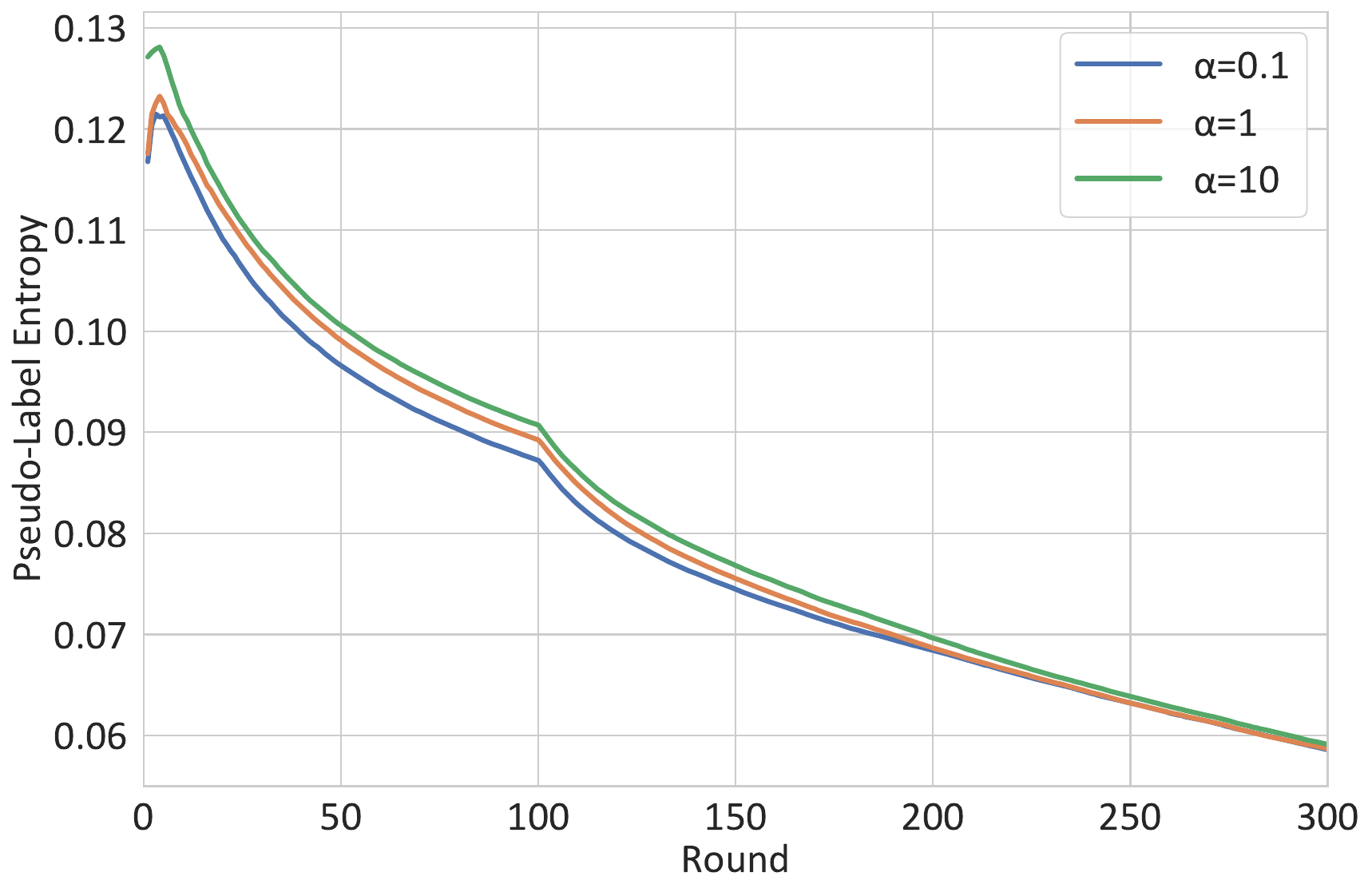}
    \caption{Pseudo-label entropy of the local model under different heterogeneity.}
  \end{subfigure}

  \caption{Changes in the pseudo-label confidence entropy of the global and local model as heterogeneity increases. Experiments show that as heterogeneity increases, global pseudo-label entropy will gradually increase, while local pseudo-label entropy will gradually decrease.}
  \label{fig:figure_entropy}
\end{figure*}

\begin{figure*}[!t]
  \centering

  \begin{subfigure}{0.24\linewidth}
    \centering
    \includegraphics[width=\linewidth]{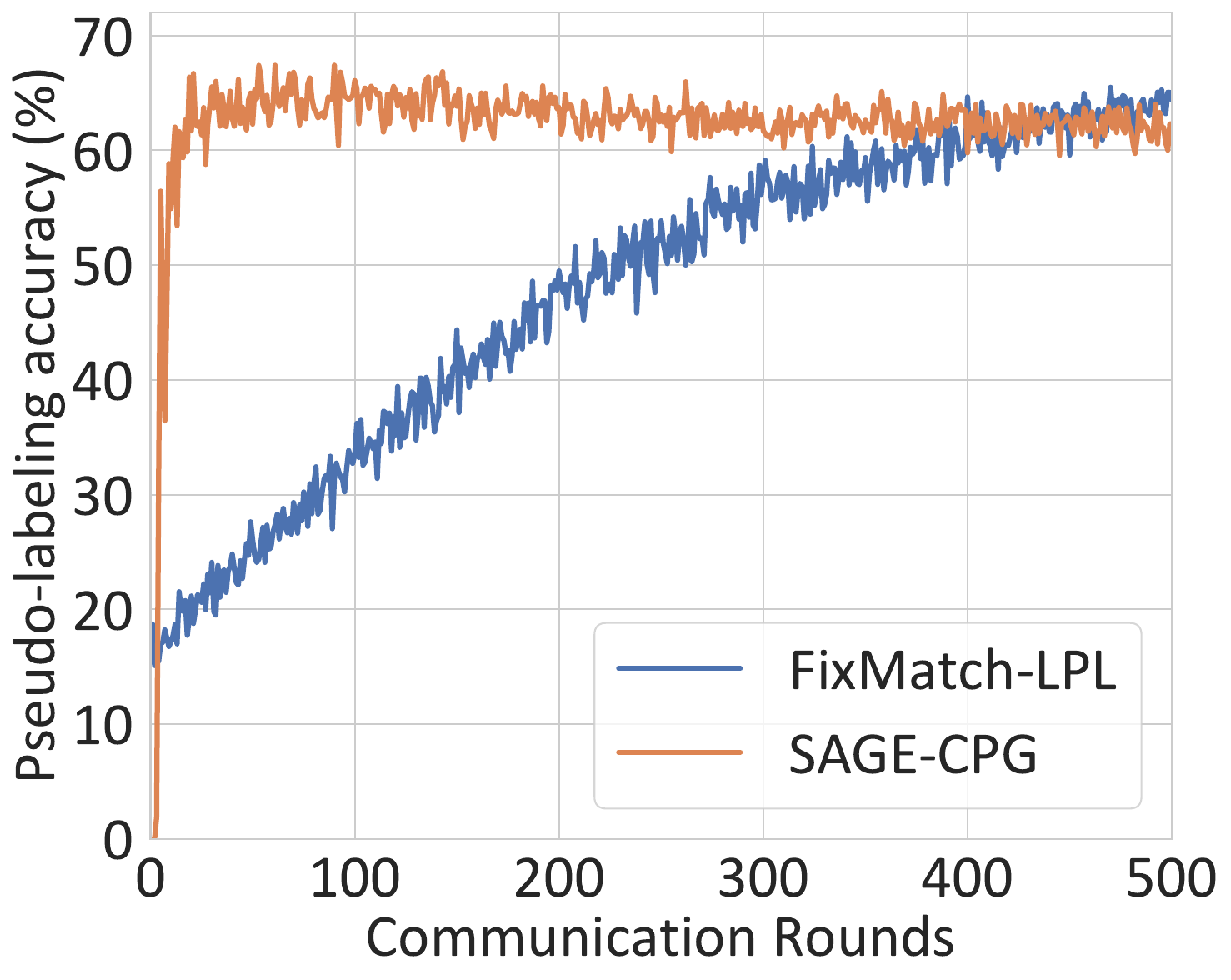}
    \caption{Pseudo-labeling accuracy with $\alpha=0.5$.}
  \end{subfigure}
  \hfill
  \begin{subfigure}{0.24\linewidth}
    \centering
    \includegraphics[width=\linewidth]{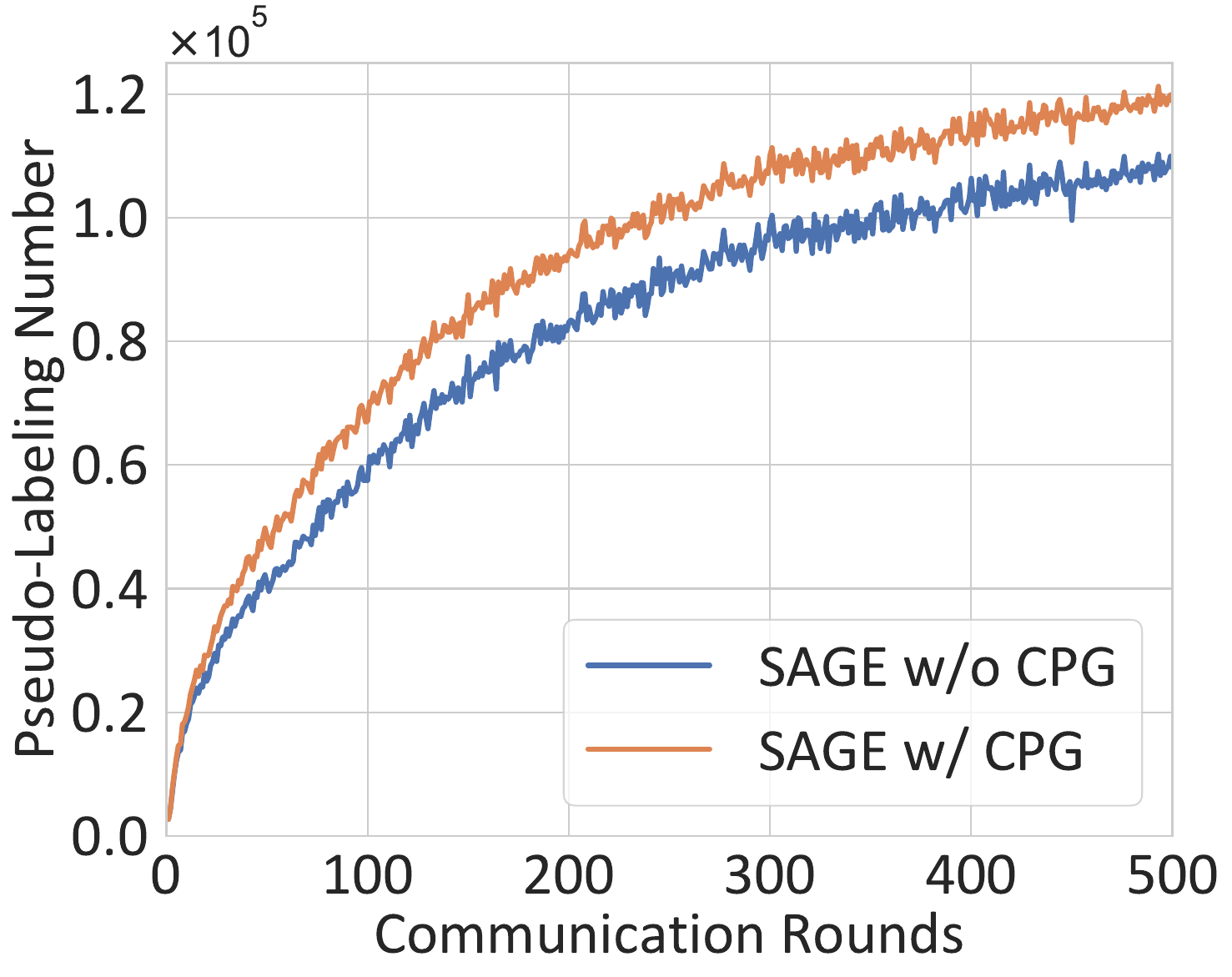}
    \caption{Comparison of the number of pseudo-labels with $\alpha=0.5$.}
  \end{subfigure}
  \hfill
  \begin{subfigure}{0.24\linewidth}
    \centering
    \includegraphics[width=\linewidth]{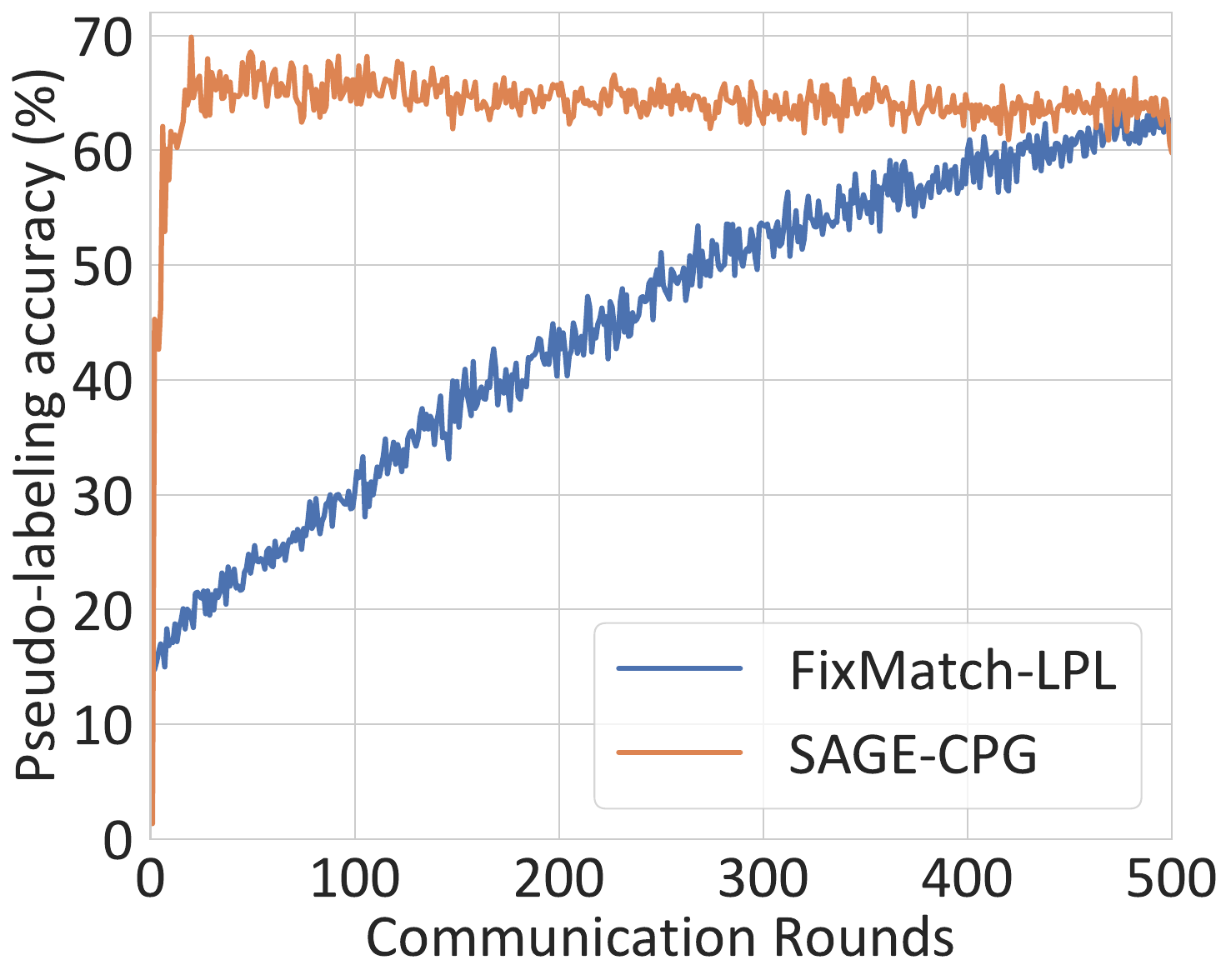}
    \caption{Pseudo-labeling accuracy with $\alpha=1$.}
  \end{subfigure}
  \hfill
  \begin{subfigure}{0.24\linewidth}
    \centering
    \includegraphics[width=\linewidth]{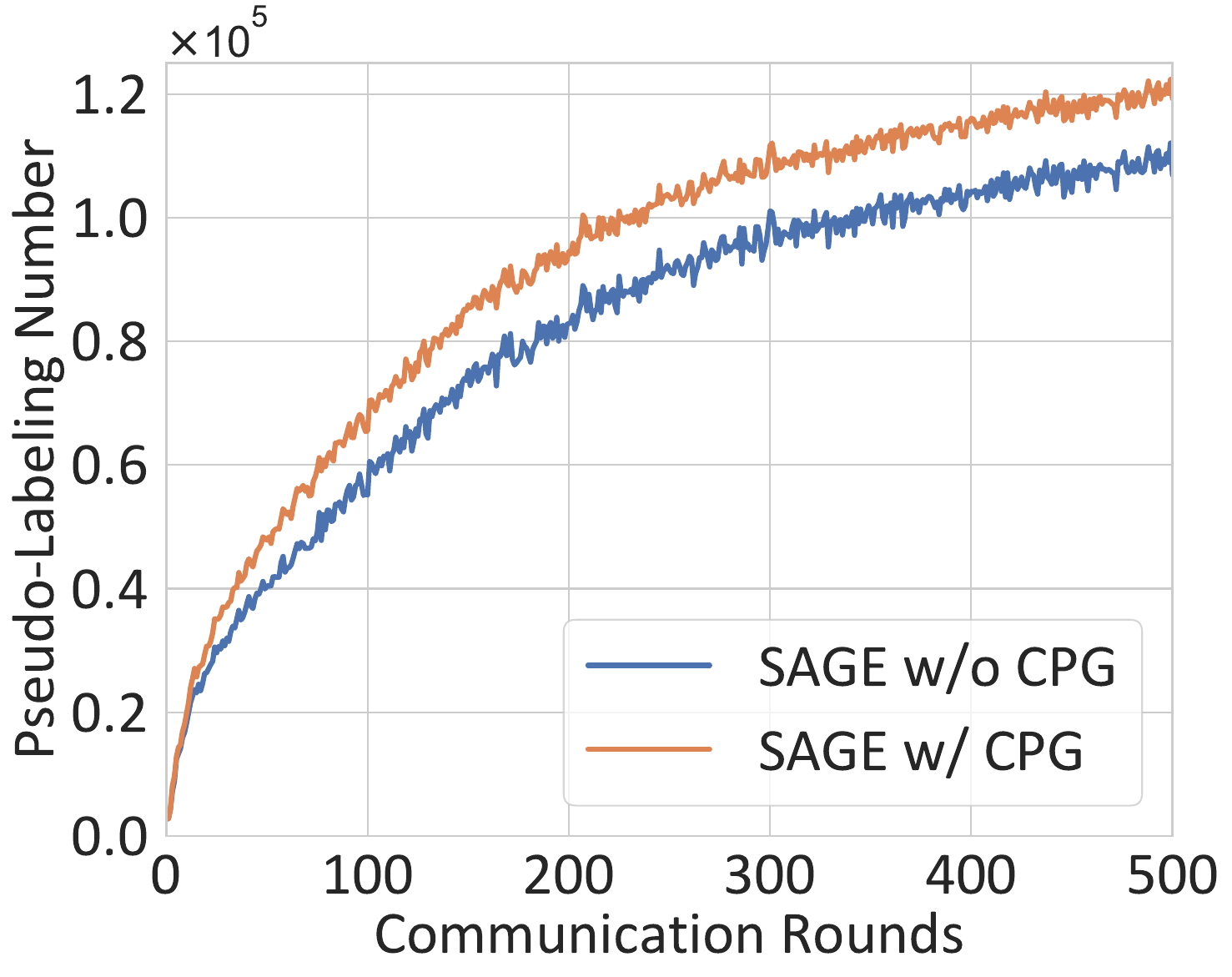}
    \caption{Comparison of the number of pseudo-labels with $\alpha=1$.}
  \end{subfigure}

  \caption{Additional ablation of CPG on CIFAR-100.}
  \label{fig:additional_cpg}
\end{figure*}

\begin{figure}[!t]
    \centering
    \includegraphics[width=\linewidth]{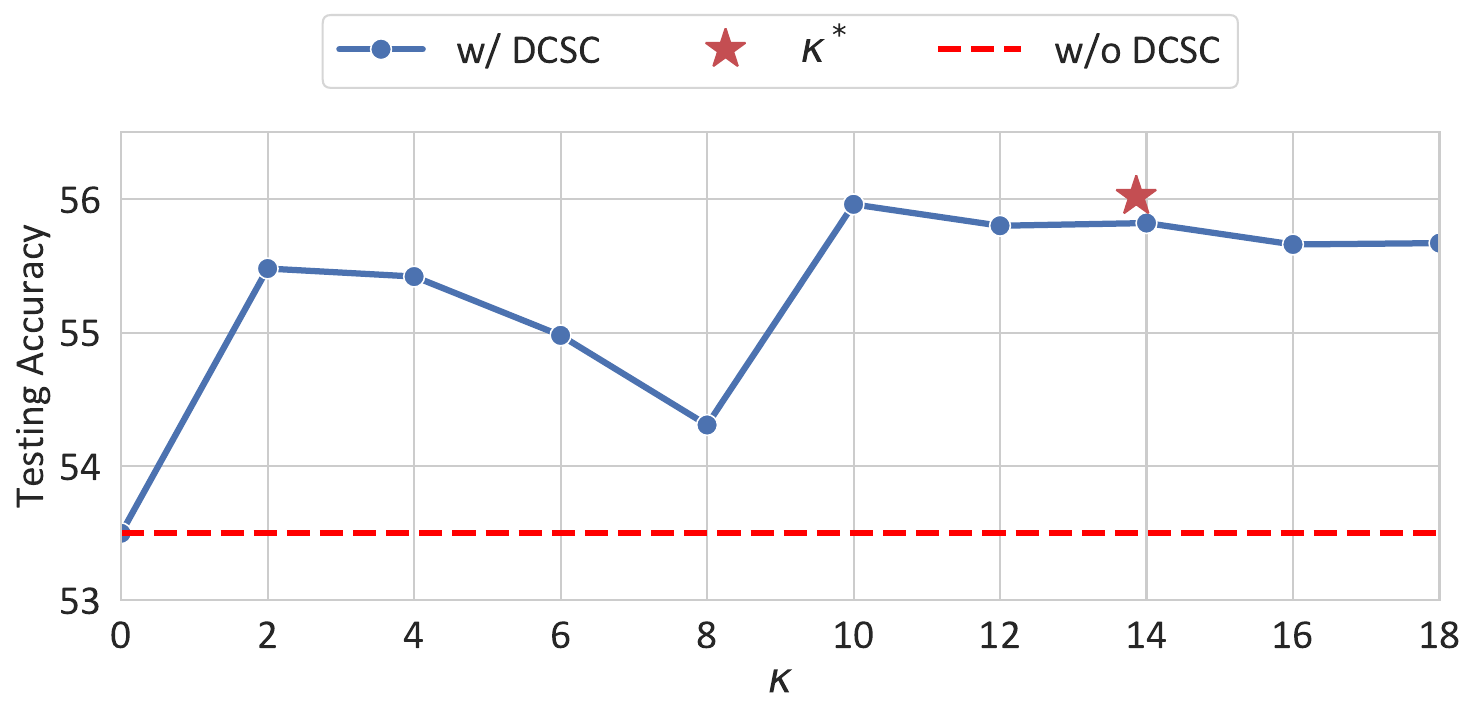}
    \caption{Ablation study on $\kappa$.}
    \label{fig:figure10}
\end{figure}

% soft label
\begin{table}[t]

\centering
  \caption{Ablation studies on soft label.}
  % \vspace{-0.1cm}
  \scriptsize
    \resizebox{1\linewidth}{!}{
\begin{tabular}{ccccc}
\hline\hline
Method                         & Label & $\alpha=0.1$   & $\alpha=0.5$   & $\alpha=1$     \\ \hline \hline
\multirow{2}{*}{FixMatch-LPL}  & Hard  & 49.32          & 49.67          & 49.55          \\
                               & Soft  & 31.96          & 33.17          & 32.61          \\ \cline{2-5} 
\multirow{2}{*}{FixMatch -GPL} & Hard  & 48.96          & 51.80          & 52.19          \\
                               & Soft  & 48.68          & 50.77          & 48.64          \\ \cline{2-5} 
\multirow{2}{*}{SAGE}          & Hard  & \textbf{54.18} & \textbf{55.82} & \textbf{56.06} \\
                               & Soft  & 53.05          & 54.53          & 55.90          \\ \hline \hline
\end{tabular}
}
  \label{table:soft_label}
\end{table}

\begin{figure}[!t]
    \centering
    \begin{subfigure}{0.48\linewidth}
        \centering
        \includegraphics[width=\linewidth]{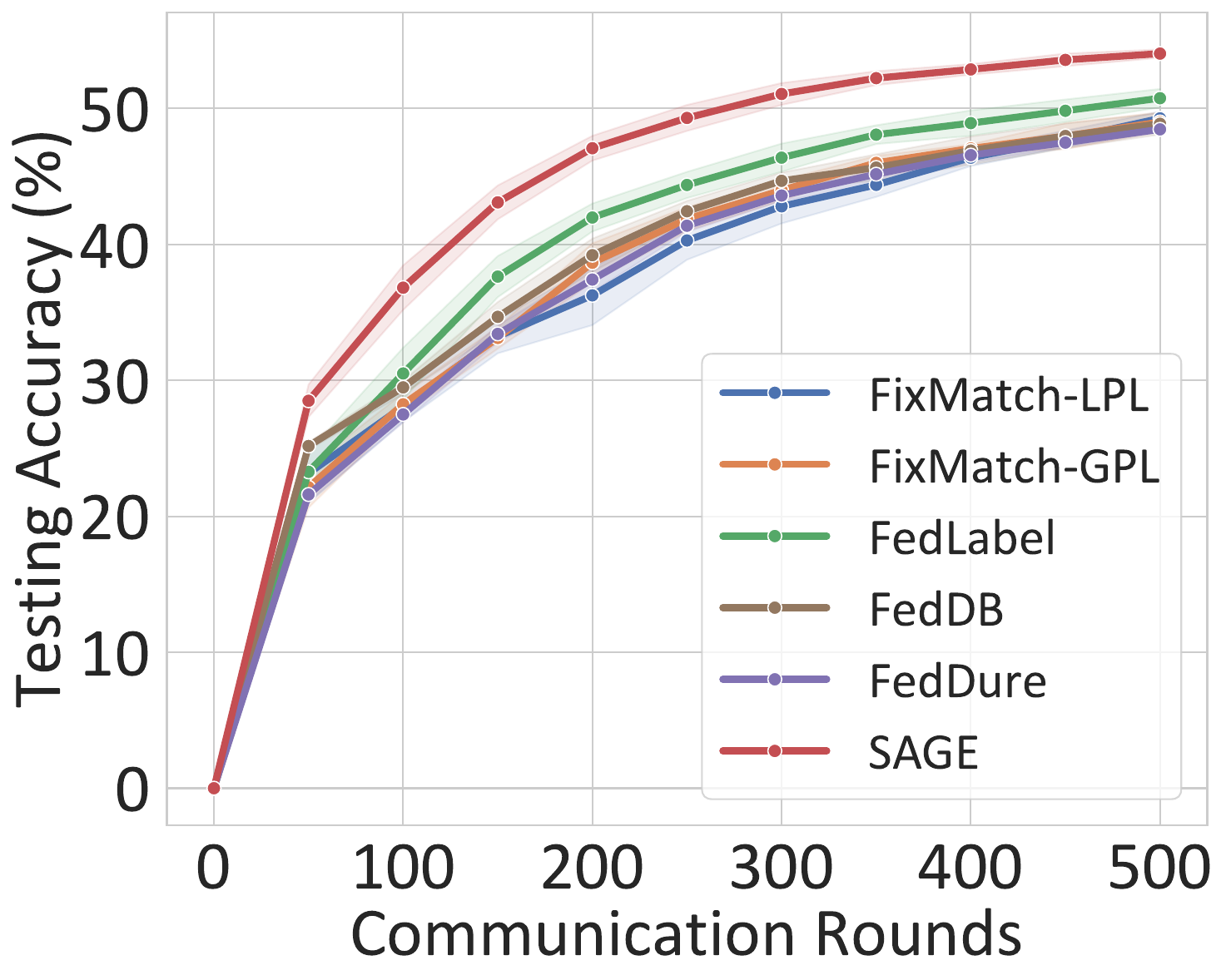}
        \caption{Convergence curves of SAGE and other baseline methods with $\alpha=0.1$.}
        \label{fig:figure11a}
    \end{subfigure}
    \hfill
    \begin{subfigure}{0.48\linewidth}
        \centering
        \includegraphics[width=\linewidth]{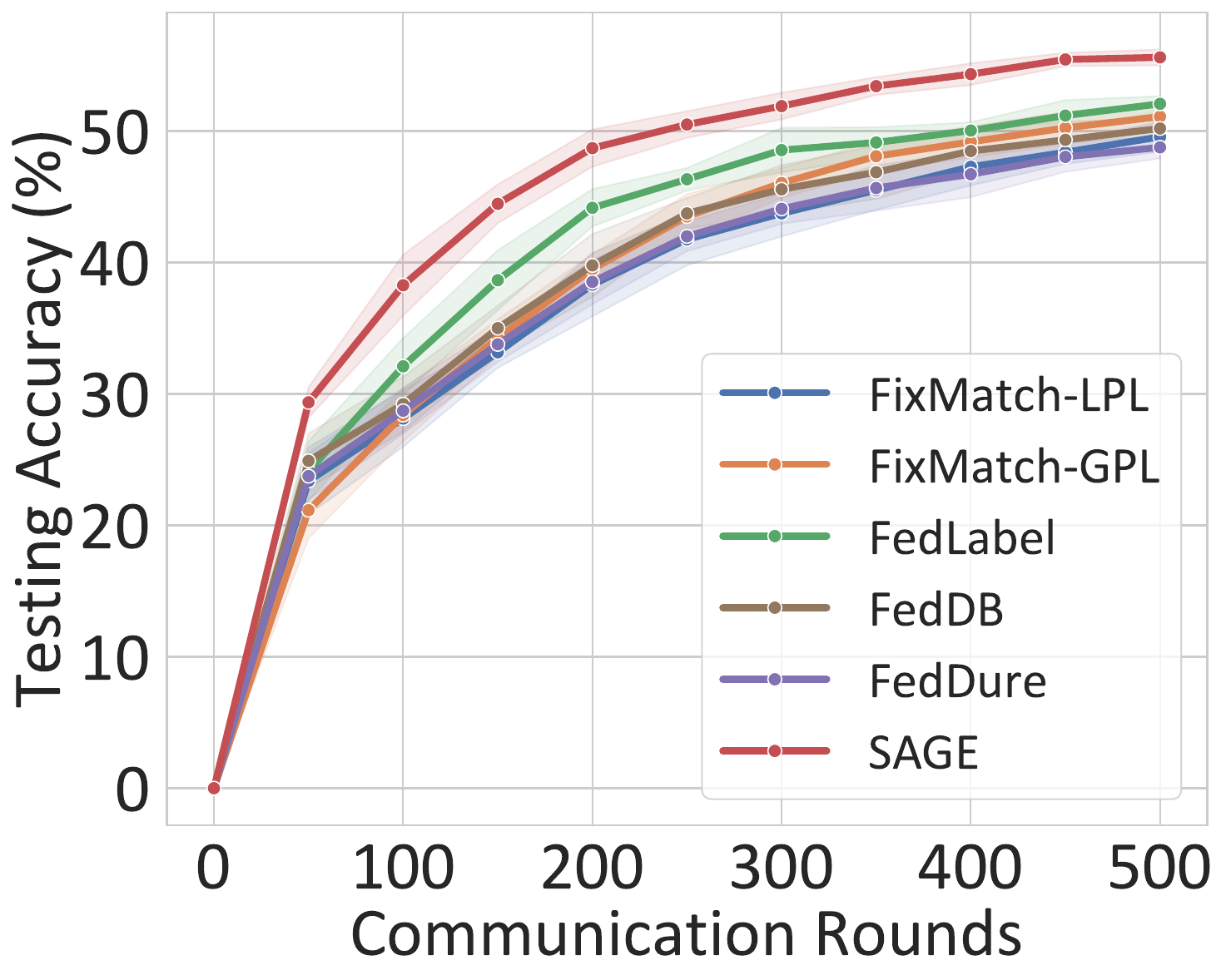}
        \caption{Convergence curves of SAGE and other baseline methods with $\alpha=0.5$.}
        \label{fig:figure11b}
    \end{subfigure}
    \caption{Additional convergence curves under different heterogeneities.}
    \label{fig:figure11}
\end{figure}

\paragraph{Local model.} For the local model, we define the entropy of the local unsupervised data distribution as $H(Q^u(y))$, aiming to explore the relationship between the entropy of the local data distribution $H(Q^u(y))$ and the entropy of model predictions $H(p(y | x, \mathcal{D}^u))$. For $p(y | \mathcal{D}^u)$, during local training, since $N^u \gg N^s$, as the local training time $t$ increases, the local model adjusts $p(y | \mathcal{D}^u)$ based on the pseudo-labels $\hat{y}_{l}^{i}$ of the unlabeled sample $\mathbf{u}$:
\begin{align}
p^{(t+1)}(y | \mathcal{D}^u) &= \gamma \cdot \left( p(\hat{y}_{l}^{i}=y | x, \mathcal{D}^u) - p^{(t)}(y | \mathcal{D}^u) \right) \nonumber \\
&\quad + p^{(t)}(y | \mathcal{D}^u).
\label{eq:local_iteration}
\end{align}
As time $t$ progresses, the prior distribution $p(y | \mathcal{D}^u)$ gradually couples with the true local unsupervised distribution $Q^u(y)$, this indicates a correlation between $H(p(y | \mathcal{D}^u))$ and $H(Q^u(y))$. For $p(y | x, \mathcal{D}^u)$, we expand it using Bayes' theorem as follows:
\begin{equation}
    p(y | x, \mathcal{D}^u) = \frac{p(x | y, \mathcal{D}^u) \cdot p(y | \mathcal{D}^u)}{p(x | \mathcal{D}^u)},
\end{equation}
here, $p(y | \mathcal{D}^u)$ denotes the prior distribution of classes, $p(x | y, \mathcal{D}^u)$ is the feature distribution, and $p(x | \mathcal{D}^u)$ is the marginal distribution.The entropy $H(p(y | x, \mathcal{D}^u))$, when expanded according to Bayes’ theorem, can be expressed as:
\begin{align}
    H(p(y | x, \mathcal{D}^u)) &= - \sum_{y} p(y | x, \mathcal{D}^u) \log p(y | x, \mathcal{D}^u) \nonumber \\
    &= - \sum_{y} \left( \frac{p(x | y, \mathcal{D}^u) \cdot p(y | \mathcal{D}^u)}{p(x | \mathcal{D}^u)} \right) \nonumber  \\
&\quad \cdot \log \left( \frac{p(x | y, \mathcal{D}^u) \cdot p(y | \mathcal{D}^u)}{p(x | \mathcal{D}^u)} \right).
\end{align}
Consider the term associated with the prior distribution $p(u | \mathcal{D}^u)$:
{\small
\begin{align}
        H(p(y | x, \mathcal{D}^u)) &=- \sum_{y} \frac{p(x | y, \mathcal{D}^u) \cdot p(y | \mathcal{D}^u)}{p(x | \mathcal{D}^u)} \log p(y | \mathcal{D}^u) \nonumber\\
        &- \sum_{y} \frac{p(x | y, \mathcal{D}^u) \cdot p(y | \mathcal{D}^u)}{p(x | \mathcal{D}^u)} \log p(x | y, \mathcal{D}^u),
\end{align}
}the first term represents the entropy of the model's prior distribution:
\begin{equation}
    H(p(y | \mathcal{D}^u)) = - \sum_{y} p(y | \mathcal{D}^u) \log p(y | \mathcal{D}^u).
\end{equation}
The second term encapsulates a component that quantifies the feature distribution:
{\small
\begin{equation}
    \text{KL}(p(x | y, \mathcal{D}^u) \parallel p(x | \mathcal{D}^u)) = \sum_{y} p(y | \mathcal{D}^u) \log \frac{p(x | y, \mathcal{D}^u)}{p(x | \mathcal{D}^u)}.
\end{equation}
}Finally, the entropy of the predictive distribution $H(p(y | x, \mathcal{D}^u))$ can be written as follows:
\begin{align}
H(p(y | x, \mathcal{D}^u)) 
&= H(p(y | \mathcal{D}^u)) \nonumber \\
&\quad + \underbrace{\text{KL}\left( p\left( x | y, \mathcal{D}^u \right) \, \bigg\| \, p\left( x | \mathcal{D}^u \right) \right)}_{\text{Contribution of features}},
\end{align}

\renewcommand{\arraystretch}{1.1} % 
% 0.1 
\begin{table*}[!t]
\centering
\caption{Comparison of convergence rates between SAGE and other baseline methods with $\alpha=0.1$.}
\begin{tabular}{ccccccccc}
\hline \hline
Acc.         & \multicolumn{2}{c}{30\%} & \multicolumn{2}{c}{40\%} & \multicolumn{2}{c}{45\%} & \multicolumn{2}{c}{50\%} \\
Method &
  Round $\downarrow$ &
  Speedup $\uparrow$ &
  Round $\downarrow$ &
  Speedup $\uparrow$ &
  Round $\downarrow$ &
  Speedup $\uparrow$ &
  Round $\downarrow$ &
  Speedup$\uparrow$ \\ \hline \hline
FixMatch-LPL & 119    & $\times$1.00    & 242    & $\times$1.00    & 360    & $\times$1.00    & 562     & $\times$1.00   \\
FixMatch-GPL & 114    & $\times$1.04    & 226    & $\times$1.07    & 322    & $\times$1.12    & 524     & $\times$1.07   \\
FedLabel     & 94     & $\times$1.27    & 175    & $\times$1.38    & 259    & $\times$1.39    & 429     & $\times$1.31   \\
FedDB        & 103    & $\times$1.16    & 206    & $\times$1.17    & 321    & $\times$1.12    & None    & None           \\
FedDure      & 114    & $\times$1.04    & 234    & $\times$1.03    & 341    & $\times$1.06    & 542     & $\times$1.04   \\
\textbf{SAGE} &
  \cellcolor[HTML]{EFEFEF}\textbf{60} &
  \cellcolor[HTML]{EFEFEF}\textbf{$\times$1.98} &
  \cellcolor[HTML]{EFEFEF}\textbf{124} &
  \cellcolor[HTML]{EFEFEF}\textbf{$\times$1.95} &
  \cellcolor[HTML]{EFEFEF}\textbf{174} &
  \cellcolor[HTML]{EFEFEF}\textbf{$\times$2.07} &
  \cellcolor[HTML]{EFEFEF}\textbf{267} &
  \cellcolor[HTML]{EFEFEF}\textbf{$\times$2.10} \\ \hline \hline
\end{tabular}
  
  \label{table:conver_0.1}
\end{table*}

This indicates that $H(p(y | x, \mathcal{D}^u))$ can be decomposed into the entropy of the prior distribution $H(p(y | \mathcal{D}^u))$ and a KL-divergence term contributed by the feature distribution. Under the heterogeneous setting, the local model struggles to establish robust feature discrimination across clients in the early stages of training, limiting the influence of the feature distribution on the predictive distribution. This implies that $H(p(y | x, \mathcal{D}^u))$ is mainly influenced by $H(p(y | \mathcal{D}^u))$, i.e., $H(p(y | x, \mathcal{D}^u)) \sim H(p(y | \mathcal{D}^u))$. Therefore, we conclude that $H(p(y | x, \mathcal{D}^u))$ is influenced by $H(p(y | \mathcal{D}^u))$ and correlates with $H(Q^u(y))$. As the degree of heterogeneity increases, $H(Q^u(y))$ decreases, consequently affecting $H(p(y | x, \mathcal{D}^u))$ and causing it to decrease accordingly.

\paragraph{Global model.} The global model updates by aggregating parameters from multiple local models, it aims to learn a ``compromise" global distribution that balances all client-side local distributions. The global model’s confidence predictions are not directly influenced by the local class distribution of any specific client. However, As the degree of non-IIDness increases, the differences between local class distributions become more pronounced. The global model cannot simultaneously satisfy the extreme requirements of each local data distribution, so it makes high-confidence predictions only for samples with greater consistency across clients: 
\begin{equation}
p(y | x, \theta_g) \approx \frac{1}{|\mathcal{C}_M|} \sum_{m=1}^{|\mathcal{C}_M|} p(y | x, \theta_{l,m}).
\end{equation}
As a result, the global model's confidence predictions increasingly focus on classes with higher consistency across clients, demonstrating more conservative prediction behavior.

% 0.5
\begin{table*}[!t]
\centering
\caption{Comparison of convergence rates between SAGE and other baseline methods with $\alpha=0.5$.}
\begin{tabular}{ccccccccc}
\hline \hline
Acc.         & \multicolumn{2}{c}{30\%} & \multicolumn{2}{c}{40\%} & \multicolumn{2}{c}{45\%} & \multicolumn{2}{c}{50\%} \\
Method &
  Round $\downarrow$ &
  Speedup $\uparrow$ &
  Round $\downarrow$ &
  Speedup $\uparrow$ &
  Round $\downarrow$ &
  Speedup $\uparrow$ &
  Round $\downarrow$ &
  Speedup$\uparrow$ \\ \hline \hline
FixMatch-LPL & 121    & $\times$1.00    & 221    & $\times$1.00    & 334    & $\times$1.00    & 546    & $\times$1.00    \\
FixMatch-GPL & 113    & $\times$1.07    & 210    & $\times$1.05    & 274    & $\times$1.22    & 419    & $\times$1.30    \\
FedLabel     & 83     & $\times$1.46    & 160    & $\times$1.38    & 222    & $\times$1.50    & 366    & $\times$1.49    \\
FedDB        & 94     & $\times$1.29    & 205    & $\times$1.08    & 282    & $\times$1.18    & 492    & $\times$1.11    \\
FedDure      & 110    & $\times$1.10    & 222    & $\times$1.00    & 315    & $\times$1.06    & 552    & $\times$0.99    \\
\textbf{SAGE} &
  \cellcolor[HTML]{EFEFEF}\textbf{55} &
  \cellcolor[HTML]{EFEFEF}\textbf{$\times$2.20} &
  \cellcolor[HTML]{EFEFEF}\textbf{105} &
  \cellcolor[HTML]{EFEFEF}\textbf{$\times$2.10} &
  \cellcolor[HTML]{EFEFEF}\textbf{159} &
  \cellcolor[HTML]{EFEFEF}\textbf{$\times$2.10} &
  \cellcolor[HTML]{EFEFEF}\textbf{241} &
  \cellcolor[HTML]{EFEFEF}\textbf{$\times$2.27} \\ \hline \hline
\end{tabular}
  
  \label{table:conver_0.5}
\end{table*}

\subsection{Experimental Support for Analysis Results}
\label{sec:experimental_support}
To support the analytical conclusions in Appendix~\ref{sec:analysis} and Remark~\hyperlink{remark1}{1} and \hyperlink{remark2}{2} in Section~\ref{section:preliminary_study}, we conducted further exploratory experiments on CIFAR-100, analyzing how the entropy of pseudo-label confidence for the local and global models changes with heterogeneity. As shown in Fig.~\ref{fig:figure_entropy}(a), when data heterogeneity intensifies, the entropy of the global model’s pseudo-label confidence tends to increase, indicating greater uncertainty. This causes the global model’s pseudo-labeling strategy to become more conservative. Conversely, in Fig.~\ref{fig:figure_entropy}(b), the entropy of the local model’s pseudo-label confidence tends to decrease as data heterogeneity increases, especially in the early stages of training when the local model has not yet developed robust feature differentiation capabilities. This suggests that the local model’s predictions become overly reliant on the local imbalanced distribution, leading to overfitting and overly confident predictions.

\section{Additional Ablation Study}
\label{section:Additional_ablation_Study}
In this section, we conduct further studies on the CPG and CDSC modules of SAGE, building on the ablation experiments in the main manuscript to demonstrate the effectiveness of these components.

\subsection{Corrected Soft Label or Direct Soft Label?}
The corrected soft labels produced by SAGE can mitigate the harmful effects of incorrect predictions. Additionally, we investigate whether directly using the model's predicted soft labels could achieve a similar effect. As shown in Tab.~\ref{table:soft_label}, directly using soft labels results in decreased performance, even worse than directly using hard labels. This is because directly using model predictions as soft labels suppresses all classes except the predicted one, thereby failing to mitigate the harm of incorrect pseudo-labels and potentially introducing extra noise. In contrast, the soft labels generated by SAGE ensure that prediction signals from both models are preserved, thereby enhancing their consensus.

\subsection{Ablation Study on the correction coefficient $\lambda$}
\label{sec:additional_ablatioin_lambda}
We define the dynamic correction coefficient $\lambda$ to regulate the contribution of local and global pseudo-labels. We conduct an in-depth study of $\lambda$ on CIFAR-100, as shown in Fig.~\ref{fig:add_figure_lambda}:
(1) According to Fig.~\ref{fig:add_figure_lambda}(a), 
$\lambda$ increases as heterogeneity intensifies, indicating that SAGE effectively detects the increase in heterogeneity and subsequently relies more on the global model. (2) According to Fig.~\ref{fig:add_figure_lambda}(b), $\lambda$ for local minority classes is smaller than that for local majority classes, suggesting that local minority classes tend to rely more on the predictions of the global model. (3) As training progresses, $\lambda$ increases, and the gap between majority and minority narrows, suggesting an increase in the consensus between the models, consistent with the conclusion in Fig.~\ref{fig:figure8}.

\subsection{Additional Ablation Study on CPG}
\label{sec:add_abl_cpg}
In Fig.~\ref{fig:figure6} of Section~\ref{sec:ablation_study}, we conducted the effectiveness analysis of CPG under the setting of $\alpha=0.1$, confirming that CPG can significantly improve the quantity and quality of pseudo-labels. In this section, we conducted additional experiments under different heterogeneity settings to verify the robustness of CPG. As shown in Fig.~\ref{fig:additional_cpg}, under the settings of $\alpha=\{0.5,1\}$, CPG is still able to generate high-accuracy pseudo-labels in the early stages of training, supplementing the local model's pseudo-label predictions for local minority classes and further enhancing the utilization of unlabeled data.

\subsection{Ablation Study on the Sensitivity Coefficient $\kappa$}
In the implementation of CDSC, $\kappa$ in Eq.~\eqref{eq:Dynamic Correction Coefficient} adjusts the sensitivity of the correction coefficient $\lambda(\mathbf{u})$ to the confidence discrepancy $\Delta C(\mathbf{u})$. On CIFAR-100, we divided clients with $\alpha=1$ and varied $\kappa$ in increments of 2 to study the robustness of SAGE with respect to $\kappa$. The results shown in Fig.~\ref{fig:figure10} indicate that CDSC remains effective regardless of the value of $\kappa$. As $\kappa$ increases, SAGE performance stabilizes, indicating low sensitivity to the hyperparameter $\kappa$.

In our experimental setup, we chose the value of $\kappa$ heuristically: we referenced the confidence interval of pseudo-labels in FixMatch, $I_{\tau}=[0.95,1]$, aiming for $\lambda(\cdot)$ to allocate equal weight to the local and global models when the confidence discrepancy reaches the interval length $|I_{\tau}|=0.05$. Thus,
\begin{equation}
    \exp(-\kappa^*\cdot |I_{\tau}|)=0.5.
\end{equation}
Solving this equation, we find $\kappa^* \approx 13.86$. In our experimental setups, $\kappa^*$ yielded the best results.

\section{Additional Comparison with Baselines}
To demonstrate the effectiveness of SAGE, we present a comparison between SAGE and baseline methods with a 10\% labeling ratio in Section 4 of the main manuscript. In this supplementary material, we further illustrate the robustness of SAGE with less or more labeled data by comparing SAGE with baseline methods at 20\% labeling ratio. Additionally, to verify that SAGE consistently improves convergence rate, we compare the convergence of SAGE and baseline methods under varying degrees of heterogeneity.

\renewcommand{\arraystretch}{1.0} % 
% 20%
\begin{table*}[!t]
  \centering
    \caption{Experimental results on CIFAR-10, CIFAR-100, SVHN and CINIC-10 under 20\% label. Bold text indicates the best result, while underlined text indicates the second-best result. The last row presents the improvement of SAGE over existing methods.}
  \resizebox{1\textwidth}{!}{
\begin{tabular}{rlcccccccccccc}
\hline \hline
\multicolumn{2}{c}{} &
  \multicolumn{3}{c}{CIFAR-10} &
  \multicolumn{3}{c}{CIFAR-100} &
  \multicolumn{3}{c}{SVHN} &
  \multicolumn{3}{c}{CINIC-10} \\
\multicolumn{2}{c}{\multirow{-2}{*}{Methods}} &
  $\alpha=0.1$ &
  $\alpha=0.5$ &
  $\alpha1$ &
  $\alpha=0.1$ &
  $\alpha=0.5$ &
  $\alpha1$ &
  $\alpha=0.1$ &
  $\alpha=0.5$ &
  $\alpha1$ &
  $\alpha=0.1$ &
  $\alpha=0.5$ &
  $\alpha1$ \\ \hline \hline
\multicolumn{2}{r}{\textbf{SL methods}} &
   &
   &
   &
   &
   &
   &
   &
   &
   &
   &
   &
   \\ \hline \hline
\multicolumn{2}{r}{FedAvg} &
  86.37 &
  87.06 &
  87.97 &
  45.72 &
  46.57 &
  47.55 &
  88.37 &
  89.05 &
  89.97 &
  66.24 &
  68.29 &
  69.21 \\
\multicolumn{2}{r}{FedProx} &
  86.78 &
  88.11 &
  88.44 &
  45.96 &
  47.33 &
  47.89 &
  87.99 &
  88.56 &
  91.10 &
  65.53 &
  69.57 &
  69.91 \\
\multicolumn{2}{r}{FedAvg-SL} &
  90.46 &
  91.24 &
  91.32 &
  67.98 &
  68.83 &
  69.10 &
  94.11 &
  94.41 &
  94.40 &
  77.82 &
  80.42 &
  81.29 \\ \hline \hline
\multicolumn{2}{r}{\textbf{SSL methods}} &
  \multicolumn{1}{l}{} &
  \multicolumn{1}{l}{} &
  \multicolumn{1}{l}{} &
  \multicolumn{1}{l}{} &
  \multicolumn{1}{l}{} &
  \multicolumn{1}{l}{} &
  \multicolumn{1}{l}{} &
  \multicolumn{1}{l}{} &
  \multicolumn{1}{l}{} &
  \multicolumn{1}{l}{} &
  \multicolumn{1}{l}{} &
  \multicolumn{1}{l}{} \\ \hline \hline
\multicolumn{2}{r}{FixMatch-LPL} &
  87.22 &
  89.61 &
  89.23 &
  56.80 &
  57.35 &
  57.59 &
  93.66 &
  94.11 &
  94.21 &
  72.51 &
  75.14 &
  76.03 \\
\multicolumn{2}{r}{FixMatch-GPL} &
  88.55 &
  {\uline{89.69}} &
  89.83 &
  57.02 &
  57.85 &
  57.85 &
  {\uline{93.89}} &
  94.12 &
  94.17 &
  76.14 &
  {\uline{77.35}} &
  {\uline{77.82}} \\
\multicolumn{2}{r}{FedProx+FixMatch} &
  87.47 &
  89.46 &
  89.56 &
  57.44 &
  57.91 &
  57.87 &
  93.60 &
  93.93 &
  94.05 &
  72.36 &
  75.15 &
  76.06 \\
\multicolumn{2}{r}{FedAvg+FlexMatch} &
  76.36 &
  78.66 &
  78.76 &
  58.24 &
  58.44 &
  58.79 &
  56.94 &
  58.58 &
  62.19 &
  73.32 &
  75.75 &
  75.95 \\ \hline \hline
\multicolumn{2}{r}{\textbf{FSSL methods}} &
  \multicolumn{1}{l}{} &
  \multicolumn{1}{l}{} &
  \multicolumn{1}{l}{} &
  \multicolumn{1}{l}{} &
  \multicolumn{1}{l}{} &
  \multicolumn{1}{l}{} &
  \multicolumn{1}{l}{} &
  \multicolumn{1}{l}{} &
  \multicolumn{1}{l}{} &
  \multicolumn{1}{l}{} &
  \multicolumn{1}{l}{} &
  \multicolumn{1}{l}{} \\ \hline \hline
\multicolumn{2}{r}{FedMatch} &
  82.44 &
  84.13 &
  85.21 &
  45.07 &
  47.29 &
  48.40 &
  93.01 &
  93.58 &
  93.76 &
  66.94 &
  68.60 &
  72.34 \\
\multicolumn{2}{r}{FedLabel} &
  87.37 &
  88.86 &
  88.93 &
  {\uline{58.63}} &
  {\uline{58.98}} &
  {\uline{59.23}} &
  93.44 &
  94.38 &
  {\uline{94.59}} &
  60.13 &
  67.30 &
  72.22 \\
\multicolumn{2}{r}{FedLoke} &
  84.57 &
  85.26 &
  86.98 &
  53.87 &
  53.67 &
  54.56 &
  93.26 &
  93.45 &
  93.57 &
  70.63 &
  71.61 &
  71.78 \\
\multicolumn{2}{r}{FedDure} &
  {\uline{88.56}} &
  89.63 &
  {\uline{89.95}} &
  56.14 &
  57.23 &
  57.89 &
  93.81 &
  {\uline{94.42}} &
  94.37 &
  {\uline{76.21}} &
  77.13 &
  77.75 \\
\multicolumn{2}{r}{FedDB} &
  85.19 &
  86.36 &
  86.65 &
  52.81 &
  54.62 &
  55.48 &
  93.22 &
  93.50 &
  94.27 &
  74.18 &
  75.00 &
  75.65 \\
\multicolumn{2}{r}{} &
  \cellcolor[HTML]{EFEFEF}\textbf{89.87} &
  \cellcolor[HTML]{EFEFEF}\textbf{90.53} &
  \cellcolor[HTML]{EFEFEF}\textbf{90.54} &
  \cellcolor[HTML]{EFEFEF}\textbf{60.86} &
  \cellcolor[HTML]{EFEFEF}\textbf{61.49} &
  \cellcolor[HTML]{EFEFEF}\textbf{62.01} &
  \cellcolor[HTML]{EFEFEF}\textbf{94.31} &
  \cellcolor[HTML]{EFEFEF}\textbf{94.56} &
  \cellcolor[HTML]{EFEFEF}\textbf{94.68} &
  \cellcolor[HTML]{EFEFEF}\textbf{77.51} &
  \cellcolor[HTML]{EFEFEF}\textbf{78.23} &
  \cellcolor[HTML]{EFEFEF}\textbf{78.77} \\
\multicolumn{2}{r}{\multirow{-2}{*}{SAGE (ours)}} &
  {\color[HTML]{009901} ↑ 1.31} &
  {\color[HTML]{009901} ↑ 0.84} &
  {\color[HTML]{009901} ↑ 0.59} &
  {\color[HTML]{009901} ↑ 2.23} &
  {\color[HTML]{009901} ↑ 2.51} &
  {\color[HTML]{009901} ↑ 2.78} &
  {\color[HTML]{009901} ↑ 0.42} &
  {\color[HTML]{009901} ↑ 0.14} &
  {\color[HTML]{009901} ↑ 0.09} &
  {\color[HTML]{009901} ↑ 1.30} &
  {\color[HTML]{009901} ↑ 0.88} &
  {\color[HTML]{009901} ↑ 0.95} \\ \hline \hline
\end{tabular}
    }

  \label{table:experimental_20per}
\end{table*}

\subsection{Convergence Rate}
\label{subsection:convergence_rate}
In Section~\ref{subsection:Convergence Rate} of the main manuscript, we conducted experiments under the $\alpha=1$ setting, where the SAGE method significantly improved model convergence speed and test accuracy on the CIFAR-100 dataset. Here, we provide a detailed comparison of SAGE and baseline performance under different heterogeneity settings. As shown in Fig.~\ref{fig:figure11}, Tab.~\ref{table:conver_0.1} and Tab.~\ref{table:conver_0.5}, SAGE still achieves substantial acceleration in early convergence speed under the settings of $\alpha=\{0.1, 0.5\}$.

\subsection{Labeling Ratio}
\label{subsection:labeling_ratio}
Tab.~\ref{table:experimental_20per} present SAGE performance compared to baseline methods at 20\% labeling ratios, respectively. SAGE consistently achieves the best performance across different labeling ratios.

\begin{figure*}[t]
    \centering
    \begin{subfigure}{0.48\textwidth}
        \centering
        \includegraphics[width=\linewidth]{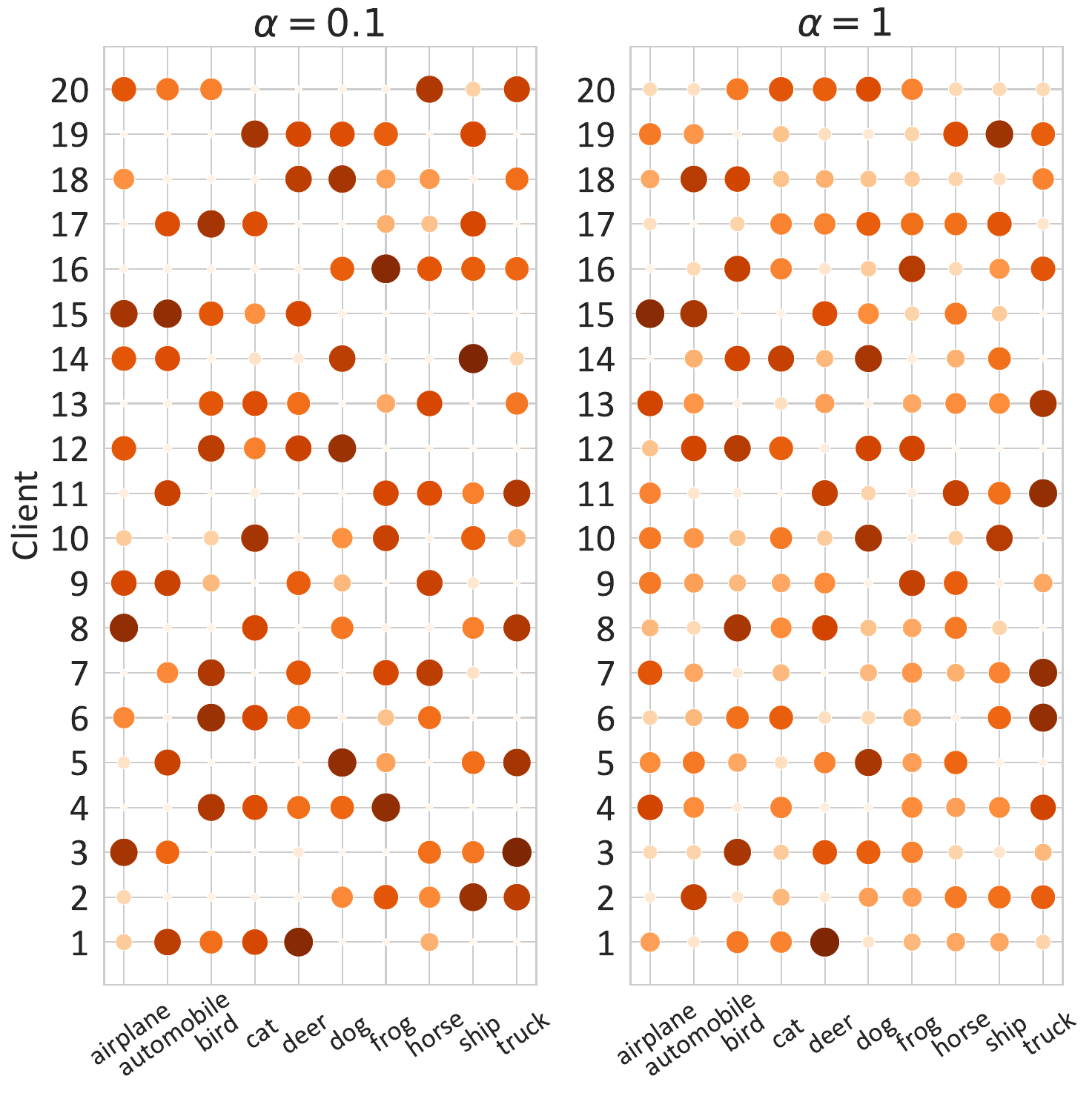}
        \caption{Labeled Distribution}
        \label{fig:figure9a}
    \end{subfigure}
    \hfill
    \begin{subfigure}{0.48\textwidth}
        \centering
        \includegraphics[width=\linewidth]{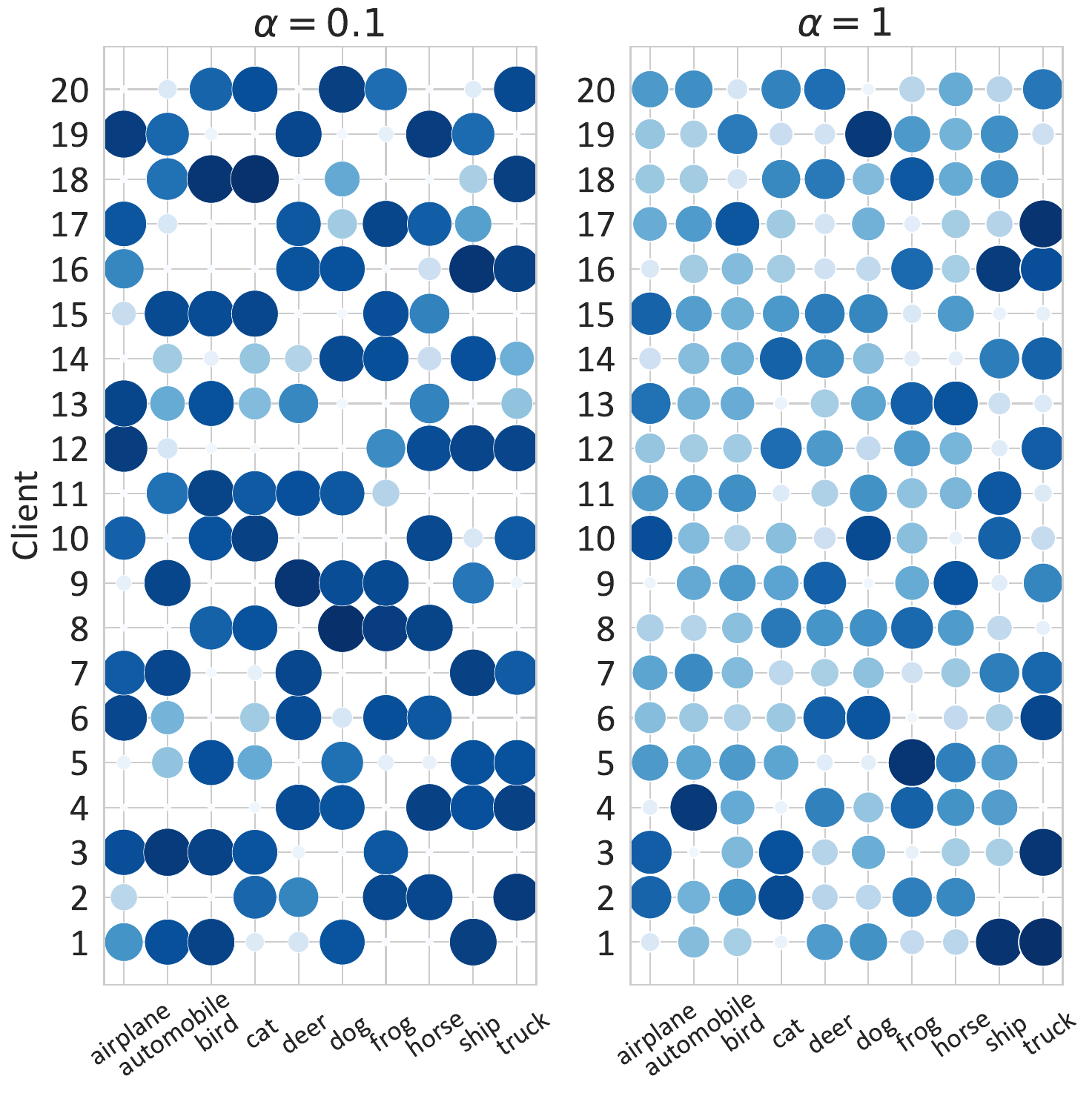}
        \caption{Unlabeled Distribution}
        \label{fig:figure9b}
    \end{subfigure}

    \caption{Distribution of labeled and unlabeled data across clients under different heterogeneity levels, using CIFAR-10 with 10\% labeling as an example. The size of each bubble represents the count of data points of class $y$ on client $k$.}
    \label{fig:figure9}
\end{figure*}

\section{Class Distribution Mismatch}
\label{sec:class_distribution_mismatch}
In this work, our experiments follow the Class Distribution Mismatch setting, where both labeled and unlabeled data within each client adhere to different heterogeneous distributions. Using CIFAR-10 as an example, Fig.~\ref{fig:figure9} shows the visualized data distribution across 20 clients.

\end{document}